\documentclass[10pt,twocolumn,letterpaper]{article}
\usepackage{rotating}
\usepackage{xspace}
\usepackage{tabularx}
\usepackage[skins]{tcolorbox}
\usepackage[table,xcdraw]{xcolor}
\newcommand{\datasetName}{{\sc MapVerse}}

\usepackage{wacv}              

\usepackage{listings}
\usepackage{tcolorbox}
\usepackage[utf8]{inputenc}
\usepackage{pifont}
\usepackage{makecell}

\definecolor{wacvblue}{rgb}{0.21,0.49,0.74}
\usepackage[pagebackref,breaklinks,colorlinks,allcolors=wacvblue]{hyperref}

\def\wacvPaperID{3261} 
\def\confName{WACV}
\def\confYear{2026}

\title{\datasetName: A Benchmark for Geospatial Question Answering on Diverse Real-World Maps}

\author{\textbf{Sharat Bhat}$^{1 *}$
\quad
\textbf{Harshita Khandelwal}$^{2}$\thanks{Equal Contribution}
\quad
\textbf{Tushar Kataria}$^{3 \dagger}$
\quad
\textbf{Vivek Gupta}$^{4}$\thanks{Equal Contribution} \\
$^{1}$ University of Southern California,
$^{2}$ University of California Los Angeles, \\
$^{3}$ University of Utah,
$^{4}$ Arizona State University\\
{\tt\small sharatpa@usc.edu,harshitaskh@ucla.edu,tushar.kataria@sci.utah.edu,vgupt140@asu.edu} 
}

\begin{document}
\maketitle
\begin{abstract}
Maps are powerful carriers of structured and contextual knowledge, encompassing geography, demographics, infrastructure, and environmental patterns. Reasoning over such knowledge requires models to integrate spatial relationships, visual cues, real-world context, and domain-specific expertise-capabilities that current large language models (LLMs) and vision–language models (VLMs) still struggle to exhibit consistently. Yet, datasets used to benchmark VLMs on map-based reasoning remain narrow in scope, restricted to specific domains, and heavily reliant on artificially generated content (outputs from LLMs or pipeline-based methods), offering limited depth for evaluating genuine geospatial reasoning.
To address this gap, we present \datasetName, a large-scale benchmark built on real-world maps. It comprises 11,837 human-authored question–answer pairs across 1,025 maps, spanning ten diverse map categories and multiple question categories for each. The dataset provides a rich setting for evaluating map reading, interpretation, and multimodal reasoning.
We evaluate ten state-of-the-art models against our benchmark to establish baselines and quantify reasoning gaps. Beyond overall performance, we conduct fine-grained categorical analyses to assess model inference across multiple dimensions and investigate the visual factors shaping reasoning outcomes. Our findings reveal that while current VLMs perform competitively on classification-style tasks, both open- and closed-source models fall short on advanced tasks requiring complex spatial reasoning. 
\end{abstract}
    
\section{Introduction}
\label{sec:intro}

\begin{figure}[h] 
    \centering
    \includegraphics[width=0.45\textwidth]{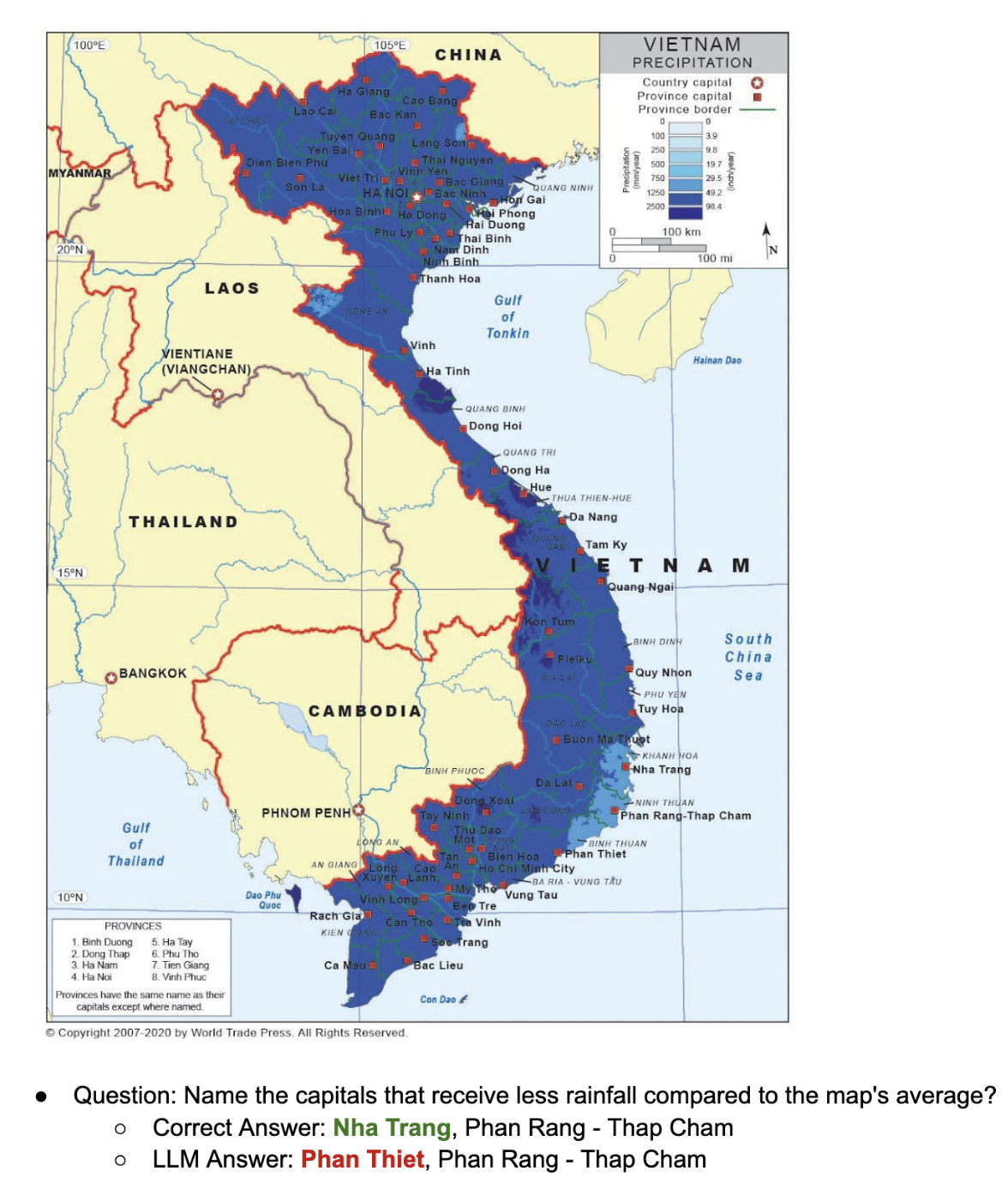} 
    \caption{\textbf{Sample in \datasetName}. A rainfall distribution map of Vietnam with a sample QA pair, showing the correct (manually annotated) answer and the predicted answer from a VLM model.}
    \label{fig:sample_Image_qa_pair}
     \vspace{-1.5em}
\end{figure}

Reasoning over visual data is central to human intelligence and has long been evaluated in cognitive tests \cite{pandya2024ntsebench,king2023administration}. The adage \textit{“a picture is worth a thousand words”} reflects how a single image can convey information rich enough for deep analysis. As LLMs and VLMs e.g., GPT \cite{achiam2023gpt}, LLaMA \cite{touvron2023llama}, and Gemini \cite{team2023gemini}, show strong gains across multimodal tasks, including mathematical problem solving \cite{pandya2024ntsebench,bandooni2025ganitbench}, scientific QA \cite{arora2023have,he2024olympiadbench}, temporal reasoning \cite{shankarampeta2025transienttables,ren2024timechat}, information synthesis \cite{khincha2023infosync,khincha2025leveraging}, and even medical and legal exams \cite{achiam2023gpt,woo2025briefme}, advanced visual reasoning emerges as a natural next frontier \cite{srivastava2022beyond,bubeck2023sparks}.
Yet research and investment still center on high-stakes domains that lend themselves to automation, such as standardized exams, code generation, clinical decision support, and autonomous driving \cite{arcadinho2024automated,schafer2023empirical,wang2024software,jain2024livecodebench,jiang2024survey,yang2024coast,zhou2025large,chen2025enhancing,singhal2023large,zhou2024vision,jiang2025alphadrive,huang2025vlm}. While impactful, this focus leaves gaps in everyday reasoning skills. A prominent omission is \textit{visual map reasoning and analysis}, crucial for navigation, planning, and spatial understanding - yet underrepresented in current benchmarks.

Maps are integral to everyday life and support a wide range of analytical and practical tasks. They enable navigation across work, travel, and leisure; delineate political boundaries and voting districts; visualize hydro-meteorological variables to study climate change and anticipate droughts or floods (Fig ~\ref{fig:sample_Image_qa_pair}); inform infrastructure planning through analyses of population movement; and guide resource management by tracking animal and fish migrations with substantial economic implications. Even in routine contexts such as finding a store in a shopping mall, maps provide compact, expressive representations that encode spatial relations, background knowledge, and domain-specific context. Reasoning over such data requires integrating spatial understanding, visual interpretation, and specialized knowledge challenges that can be difficult even for humans \cite{mukhopadhyay2025mapwise,srivastava2025apiq}. Consequently, maps serve as a powerful testbed for evaluating complex reasoning capabilities of LLMs and VLMs. However, existing benchmarks for visual reasoning over maps remain narrow in scope: they include relatively few questions \cite{mukhopadhyay2025mapwise,srivastava2025apiq}, span only a small set of domains with at most three map categories \cite{li2025mapqa,mukhopadhyay2025mapwise,srivastava2025apiq}, and rely heavily on LLM-generated or pipeline-synthesized content with minimal human annotation \cite{li2025mapqa,mukhopadhyay2025mapwise,srivastava2025apiq}. These design choices constrain question formats and limit coverage of real cartographic variability (such as legends, symbology, scales, and geographic granularity), yielding datasets that are too shallow to rigorously assess true map-centric reasoning.

To close this gap, we introduce \datasetName, a large-scale benchmark built from diverse, real-world maps paired with human-authored questions that integrate geographic reasoning, visual interpretation, and structured language understanding. The dataset contains 11,837 question–answer pairs across 1,025 maps, spanning 10 categories with multiple question types per map. It supports varied answer formats—single-entity, boolean, multi-step reasoning, ranking, counting, and open-ended—enabling rigorous evaluation of map reading and multimodal understanding. Unlike prior efforts (MapQA \cite{chang2022mapqa}, MAPWise \cite{mukhopadhyay2025mapwise}, MapIQ \cite{srivastava2025apiq}), \datasetName ~broadens the range of map types and question styles, introduces a comprehensive taxonomy such as map categories, reasoning types, geographic granularity, question formats. We conduct systematic evaluations of state-of-the-art VLMs and group-wise analyses that surface failure modes unique to map understanding at scale, establishing strong baselines and charting concrete directions for advancing geospatial and visual reasoning.  Our main contributions are:
\begin{itemize}
    \item \datasetName:  A large-scale VQA dataset comprising about 11,837 human-authored question–answer pairs over 1025 maps, covering ten diverse map categories and a wide range of question types.
    \item Extensive benchmarking of SOTA closed- and open-source VLMs, providing a detailed analysis of their current limitations. 
\end{itemize}
The \datasetName~ dataset along with associated code can be found here: \href{https://coral-lab-asu.github.io/mapverse}{https://coral-lab-asu.github.io/mapverse}.

\section{Related Works}
\textbf{Visual Question Answering (VQA) as a Lens for Evaluating Complex Reasoning Capabilities in VLMs}. The Visual Question Answering (VQA) task, introduced by Antol et al. (2015), highlighted the need for models that can jointly reason over visual and textual inputs. Core VQA skills include object recognition, attribute identification, counting, and spatial reasoning, as demonstrated in VQA v1/v2 \cite{antol2015vqa,goyal2017making}. 
Higher-order reasoning can be evaluated using datasets such as CLEVR \cite{johnson2017clevr} for compositional reasoning, Action Genome \cite{ji2020action} for action-centric inference, VCR \cite{zellers2019recognition} for commonsense reasoning, and NTSEBench \cite{pandya2024ntsebench} for cognitive reasoning skills.
More complex reasoning often requires multi-step VLM/LLM inferences, as exemplified by GQA \cite{hudson2019gqa} with reasoning chains, TallyQA \cite{acharya2019tallyqa} for counting and listing, TransientTables \cite{shankarampeta2025transienttables} for temporal reasoning, and VQARank \cite{chen2024vqarank} for ranking-based answers. Analyzing charts, maps, and diagrams similarly requires integrating multiple reasoning skills, as these visualizations condense complex data into compact, information-rich formats. Existing benchmarks include DVQA \cite{kafle2018dvqa} and PlotQA \cite{methani2020plotqa}, which emphasize structured numerical reasoning, and MapQA \cite{chang2022mapqa} and MAPWise \cite{mukhopadhyay2025mapwise}, which focus on spatial and causal reasoning over geographic data. 
Interpreting such visualizations requires domain knowledge, awareness of data-collection context, and fluency with representational conventions - bar charts, heatmaps, choropleths, and network diagrams. Without these complexities, reliable insight is hard to extract, underscoring the need for models that support robust, multidimensional visual reasoning. Motivated by this, we introduce a map-based VQA dataset that unifies these dimensions and enables rigorous evaluation of state-of-the-art VLMs across map categories, geographic granularity, and open- vs. closed-ended questions, while exposing limitations and guiding future work.

\begin{table*}[!htb]
\small
\centering
\setlength{\tabcolsep}{4pt}
 \begin{tabular}{p{1.5cm} p{2.5cm} p{3.0cm} p{9.1cm}}
\textbf{Paper} & \textbf{Size} & \textbf{Map Categories} & \textbf{Key Dataset Charactertics} \\
\midrule
\textbf{MapQA} & $\sim$800K / 60K maps & Choropleth & A collection of real, regenerated, and synthetic maps.  Synthetic QA pairs generated via a language-based pipeline.
\\

\textbf{MAPWise} & 3K / 100 maps & Choropleth  & No Real-World Maps, but Maps Generated from Consensus Datasets Paired with Human-Annotated QA pairs\\
\textbf{MapIQ} & 14.7K / 774 maps & Choropleth, Cartogram, Proportional &  Pipeline-generated maps with template-based generated QA. \\
\textbf{PEACE} & ~3.1k / 124 maps & Map Body, Legends, Stratigraphic Column, Cross Sections & 
Global geological survey maps, paired with expert-crafted QA spanning extraction, grounding, and reasoning tasks across map components, supplemented by a geology-aware agent for holistic interpretation.
\\
\datasetName & 11.8K / 1,025 maps & 10 diverse types (Refer Table \ref{tab:map_types_compact_example}) &  Real-World Maps Scraped from the Web, with Human-Generated QA Pairs \\
\bottomrule
\end{tabular}
\vspace{-0.5em}
\caption{\textbf{Comparison of Map VQA Benchmark Datasets.} Size, scope, goals, and key-characteristics for MapQA~\cite{li2025mapqa}, MAPWise~\cite{mukhopadhyay2025mapwise}, MapIQ~\cite{srivastava2025apiq}, \textbf{PEACE}~\cite{huang2025peace} and our proposed \datasetName.}
\label{tab:mapqa_comparison}
\vspace{-1.5em}
\end{table*}

\textbf{Map-Based VQA.} Recent efforts to evaluate VLM capabilities on maps include MapQA \cite{chang2022mapqa}, which introduced a large-scale benchmark of choropleth maps with V-MODEQA for table-based QA; MAPWise \cite{mukhopadhyay2025mapwise}, which evaluated maps from three regions using prompting strategies and counterfactual analysis; and MapIQ \cite{srivastava2025apiq}, which created 14.7K QA pairs across three map categories. While these benchmarks provide valuable insights, they rely heavily on synthetic content generated via LLMs or pipelines, cover a limited range of map types, and restrict question formats. Our proposed \datasetName ~addresses these limitations by offering a comprehensive benchmark of diverse, real-world maps with human-annotated questions that integrate geographic reasoning, visual interpretation, and structured language understanding. Key distinctions relative to existing benchmarks are summarized in Table \ref{tab:mapqa_comparison}. We further introduce a fine-grained taxonomy covering map categories, reasoning types, geographic granularity, and question formats, and perform extensive evaluations of state-of-the-art VLMs, systematically revealing the unique challenges of map-centric visual question answering, which is also absent from other benchmarks. By providing both scale and diversity, \datasetName ~lays the foundation for advancing research in robust, real-world map reasoning.

\section{\datasetName} \label{sec:dataset}

\textbf{Dataset Collection Process.} To ensure diversity and realism, we curated maps from publicly available sources across the internet, including news portals, educational resources, weather forecast platforms, and other open forums. Targeted keyword searches were performed for each map type, yielding an initial pool of candidates (refer to supp. \ref{ssec:web_searches}). Maps that were blurred, cropped, noisy, or unclear were discarded, retaining only high-quality and representative samples. The final dataset comprises \textit{1,025} maps spanning \textit{ten distinct map categories}, carefully selected to capture both thematic and structural diversity. Table~\ref{tab:map_types_compact_example} summarizes the distribution across different categories. Additional details on the collection and annotation process are provided in the Supplementary.

\begin{table*}[!htb]
\small
\centering
\setlength{\tabcolsep}{5pt}
\begin{tabular}{@{}l c l l@{}}
\textbf{Map Type} & \textbf{Count} & \textbf{Description} & \textbf{Example} \\
\midrule
Network       & 177 & Interconnected nodes/edges showing relations or flows. & Subway map, Human migration map \\
Divisional    & 175 & Administrative or political boundaries. & US state map, Indian district map \\
Choropleth    & 167 & Areas shaded proportionally to a quantitative variable. & Population Density, Election result map \\
Layout        & 101 & Schematic diagrams emphasizing organization. & Campus layout, Building floor plan \\
Marker        & 92  & Symbols/icons marking specific locations or data points. & Restaurant locations, Weather stations \\
Mixed         & 90  & Combines multiple map types in single representation layer. & Tourist map with POIs, Marker+Route \\
Isopleth      & 79  & Continuous contour lines connecting points of equal value. & Temperature map, Elevation map \\
Multiple      & 60  & Sets of maps for comparison or temporal data. & Flood effect maps, Yearly climate maps \\
Conflict      & 45  & Depicts tension, warfare, or territorial disputes. & War zone map, Territory change map \\
Cartogram     & 39  & Distorted regions reflecting variable magnitudes. & Population cartogram, GDP cartogram \\
\midrule
\textbf{Total} & \textbf{1,025} & & \\
\bottomrule
\end{tabular}
\vspace{-0.5em}
\caption{\textbf{Distribution and Definitions of Map Categories in \datasetName.}}
\label{tab:map_types_compact_example}
\vspace{-1.0em}
\end{table*}
\vspace{-1em}
\paragraph{Question$-$Answer (QA) Generation and Validation.} A total of \textbf{11,837 (QA) pairs} were produced using Amazon Mechanical Turk (AMT) with over 50 annotators recruited from graduate-level backgrounds in computer science, NLP, and geography. Annotators followed detailed instructions explicitly designed to encourage diverse questioning styles (refer to supp. section \ref{ssec:annotator_instructions}). Unlike benchmarks that post-edit annotations for grammar or style, we deliberately preserved natural linguistic variations and spelling errors as long as it is human intelligible, to simulate real-world scenarios, requiring models to handle realistic, noisy inputs. Each QA pair underwent independent verification by two annotators as part of a quality control process to ensure clarity, accuracy, and consistency. 

To enable detailed analysis and taxonomy, each map and its associated QA pairs were enriched with extensive metadata. For every collected map image, three types of metadata were annotated: (1) \textbf{Map Type.} Each map is assigned to one of ten categories, reflecting both structural and thematic diversity. Table~\ref{tab:map_types_compact_example} reports the count, description, and examples for each category. As shown, \datasetName ~exhibits high diversity and reasonable balance, with cartogram and conflict maps being the least represented. (2) \textbf{Image Resolution.} Image resolution plays a critical role in visual reasoning questions, as higher resolutions provide greater object clarity and potentially higher accuracy. To study this effect, we annotated five resolution ranges, from low-resolution ($<$0.1MP) to very high-resolution ($>$8MP). As shown in Table~\ref{tab:image_size_distribution}, very low- and very high-resolution images are relatively rare compared to medium-resolution images. (3) \textbf{Geographic Granularity.} Maps vary in their geographical context, ranging from localized levels such as building, town, or city, to broader scales covering countries, continents, and the world for studies like climate change. Accordingly, each map is annotated with one of twelve geographic levels, enabling evaluation across spatial hierarchies. As shown in Table~\ref{tab:geographic_level_distribution}, the dataset has fewer maps at the county, district, and state levels, and a higher number at the country and world levels.

\begin{table}[ht]
\small
    \centering
    \begin{tabular}{lr}
        \toprule
        \textbf{Question Reasoning Class} & \textbf{Count} \\
        \midrule
        Spatial        & 2,524 \\
        Visual        & 2,239 \\
        Comparative        & 1,021 \\
        Visual,Spatial     & 1,187 \\
        Spatial, Comparative     & 978   \\
        Visual, Comparative     & 802   \\
        Visual, Spatial, Comparative  & 549   \\
        Other    & 2,537 \\
        \bottomrule
    \end{tabular}
    \vspace{-0.5em}
    \caption{\textbf{Distribution of Question Reasoning Classes.}}
    \label{tab:question_aspect_types_compact}
    \vspace{-1.5em}
\end{table}

For each question, the following metadata was collected:
\begin{itemize}[leftmargin=*]
    \item \textbf{Reasoning Classes.} Based on the reasoning skills required to arrive at the correct answer, we classified each QA pair into the following categories:
\begin{itemize}[leftmargin=*]
    \item \textbf{Visual Reasoning.} Involves interpreting colors, shades, patterns, symbols, iconography, and overall visual appearance on the map.  
    \item \textbf{Spatial Reasoning.} Focuses on positions, distances, directions, and spatial relationships between regions.  
    \item \textbf{Comparative Reasoning.} Requires evaluating relative attributes or values between regions, objects, or data points, such as ranking or identifying differences.  
    \item \textbf{Other.} Includes questions that do not clearly belong to Visual, Spatial, or Comparative categories, such as metadata-related, contextual, or ambiguous questions. These questions are labeled exclusively as “Other” and are not combined with the other reasoning types.  
\end{itemize}
Questions can belong to multiple categories (Visual, Spatial, and/or Comparative) simultaneously to capture real-world complexity, except for those labeled as 'Other.' Table~\ref{tab:question_aspect_types_compact} shows the count of questions in each reasoning class.

    \item \textbf{Answer Formats.} Each question’s answer can be classified into one of six formats: Boolean, Single Entity, Counting, Listing, Ranking, and Reasoning. This categorization captures the diversity of responses, from simple presence checks to multi-step inferences or structured outputs. Table~\ref{tab:answer_types_compact} defines each format and provides their counts. Notably, \datasetName ~contains a large number of single-entity questions, while reasoning questions are the least frequent.

\end{itemize}

\begin{table}[htb]
\small
\centering
\setlength{\tabcolsep}{2pt}
 \begin{tabular}{p{1.4cm} p{0.9cm} p{5.3cm}}
\toprule
\textbf{Answer Format} & \textbf{Count} & \textbf{Description} \\
\midrule
Single Entity & 5575 & Single location, label, or map entity (e.g., state, city, or point of interest). \\
Counting      & 2395 & Count entities or occurrences (e.g., number of rivers or cities). \\
Boolean       & 1982 & Yes/no or true/false based on presence, condition, or attribute. \\
Listing       & 927  & List multiple entities satisfying a condition (e.g., all countries in a region). \\
Ranking       & 654  & Order entities by quantitative or qualitative attribute (e.g., left to right). \\
Reasoning     & 304  & Multi-step inference or contextual reasoning, final output structured. \\
\bottomrule
\end{tabular}

\caption{\textbf{Distribution of Answer Format and their definitions.} Distribution of answers formats across \datasetName.}
\label{tab:answer_types_compact}

\end{table}

\begin{figure*}[!htb]
    \centering
    \includegraphics[width=0.85\linewidth]{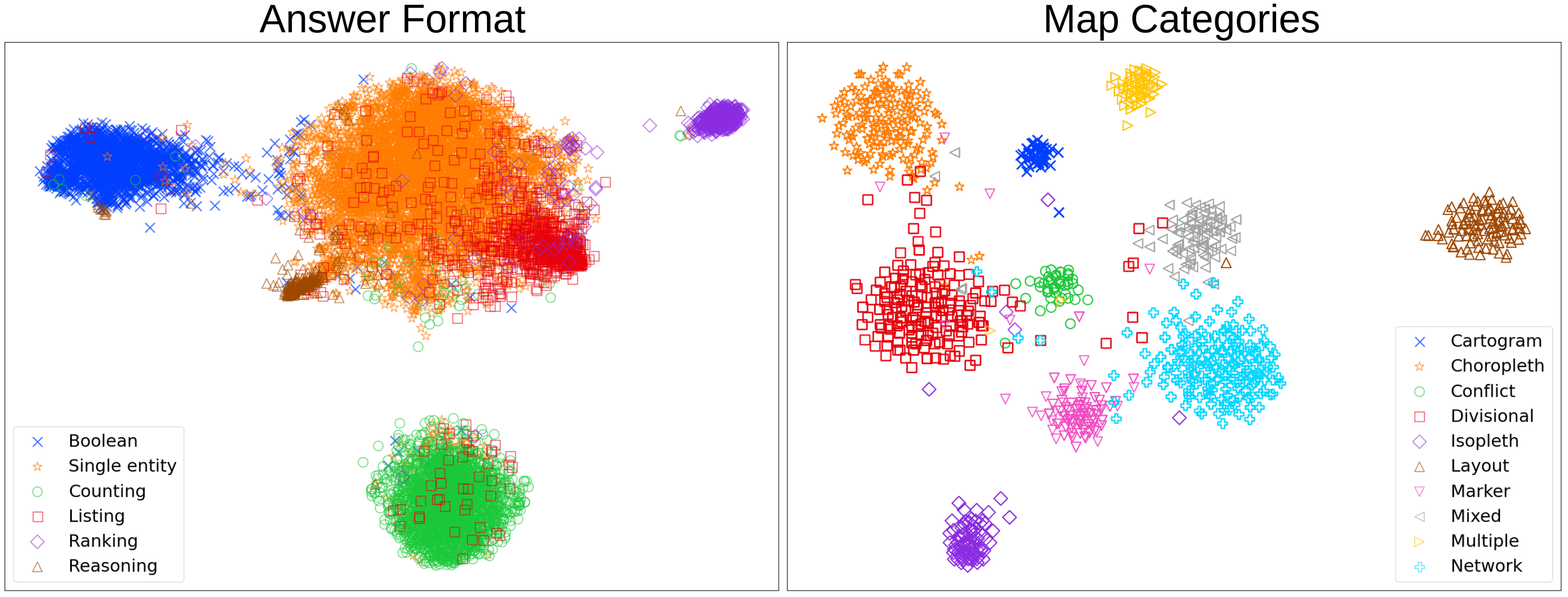}
    \caption{\textbf{t-SNE plots of \datasetName ~for different Answer Formats and Map Categories.} We used ResNet-18 encoder to obtain the latent representation of the images for different map categories and used Sentence Transformer to obtain the latent representation of the questions belonging to different answer formats. t-SNE plots were computed with a perplexity of 800 to maximize inter-class separation.}
    \label{fig:t-sne}
\end{figure*}

\noindent \textbf{Latent Space Image and QA Distribution}. From the metadata analysis above, it is evident that \datasetName ~is highly diverse across multiple aspects, including diverse map categories, reasoning classes, geographic granularity, and answer formats. However, an important dimension of diversity lies in the latent (embedding) space of images and questions. To examine this, we generated t-SNE plots of map images and answers to assess whether diversity in the aforementioned characteristics translates into diversity in the feature space of deep learning models. Figure \ref{fig:t-sne} shows t-SNE visualizations of the dataset, illustrating the distribution of maps across categories and QA pairs across answer formats. We observe that all map categories exhibit a highly diverse feature space distribution. In contrast, this is not the case for answer formats: List-type answers almost merge with Single Entity answers, suggesting that the underlying structure of questions for these two types is similar, with the only difference being multiple valid outputs for the former versus a single output for the latter. The t-SNE plot illustrating geographic granularity is shown in supp figure \ref{fig:tsne-geographical-level}.

\section{Models and Evaluations} \label{sec:benchmark}

We benchmark vision-language models (VLMs) on a map-based question-answering task (Q→A), where the model receives a map and a question about the map and produces a structured answer. To assess visual grounding independently of textual memorization, we evaluate each model under two input settings:(1) \textbf{Image + Text.} The full map image is presented alongside the question.(2) \textbf{Text Only.} Only the question is provided, with no visual input. Since the maps are sourced from publicly available real-world data, some may have been encountered during large scale pre-training prevalent in both closed and open source VLMs. In such cases, models could rely on memorized associations rather than true visual reasoning—a phenomenon we refer to as \textit{muscle memory}. By comparing performance across the two settings, we can isolate the contribution of visual input and evaluate whether models are reasoning from the image itself or relying on prior exposure.

\vspace{-1em}
\paragraph{Evaluation Metrics.} Given the variety of answer formats in our dataset, we define separate evaluation metrics for each format. This leads to multiple metrics, reflecting the diverse nature of \datasetName.

\begin{itemize}[leftmargin=*]
    \item \textbf{Boolean.} Accuracy, reflecting correct binary responses.
    \item \textbf{Counting.} Exact Match (EM), requiring predicted counts to match ground truth exactly.
    \item \textbf{Single-Entity.} Exact Match (EM), as there is only one correct entity per question.
    \item \textbf{Listing.} Precision and Recall, to measure the correctness and completeness of returned sets.
    \item \textbf{Ranking.} Mean Rank-wise Precision (RWP), assessing how well the predicted ordering matches the ground-truth at every recall stage.
    \item \textbf{Reasoning.} Exact Match (EM), requiring precise alignment between prediction and ground-truth answer.
\end{itemize}
This careful pairing of metrics with answer format enables fine-grained evaluation of model capabilities, highlighting strengths and weaknesses in visual grounding, spatial reasoning, and multi-step inference.

\vspace{-1em}
\paragraph{VLM Models Used.} We benchmark \datasetName ~using a diverse set of state-of-the-art (SOTA) open- and closed-source models, evaluating performance across a variety of architectures: \textbf{Closed-Source Models:} GPT 4o \cite{openai20244o} and  Gemini 2.5 \cite{comanici2025gemini}, which are proprietary models with advanced multimodal features. \textbf{Open-Source Models:} Aya-Vision-8B \cite{dash2025aya}, CogVLM2-19B \cite{hong2024cogvlm2},  DeepSeek-VL2-Small \cite{wu2024deepseek}, Idefics3-8B \cite{laurenccon2024building}, InternVL3-8B \cite{zhu2025internvl3}, Llama-3.2-11B-Vision-Instruct \cite{grattafiori2024llama}, Mistral-Small-3.2-24B-Instruct-2506 \cite{jiang2023clip}, Molmo-7B \cite{deitke2025molmo}, and Qwen-2.5-VL-7B-Instruct \cite{bai2025qwen25vltechnicalreport}. These models are accessible, represent diverse architectural designs, and contribute to reproducible research in the community.

\vspace{-1em}
\paragraph{Prompting Strategy.} To ensure consistency and minimize prompt-induced bias, we adopt a standardized instruction-style prompting approach. A single, fixed prompt is applied to all map-question pairs, regardless of question type or input modality. Each prompt specifies the reasoning process and guides models to produce outputs in a structured format suitable for evaluation.  Example prompt is shown in Supplementary section \ref{ssec:prompt}. The prompt includes a few-shot example Q\&A pairs covering all answer types (Boolean, Single-Entity, Counting, Listing, Ranking, Reasoning) without revealing the type of any evaluation question. This ensures that models understand the expected structure of the responses while avoiding task-specific cues. All examples are disjoint from the evaluation set, preventing data leakage.

\section{Results and Discussion}
With the experiments and discussion we want to answer the following questions:
\begin{itemize}
    \item \textit{Is \datasetName ~a challenging benchmark for current open- and closed-source state-of-the-art VLMs?}
    \item \textit{How do current VLMs perform across different map categories and question types?}
    \item \textit{Are VLMs robust to changes in map image resolution or to perturbations caused by noisy captures?}
\end{itemize}

\paragraph{\datasetName ~is a challenging dataset for currently SOTA VLMs.} Table \ref{tab:answer-type} presents results for all open-source and closed-source models, grouped by answer format due to differing evaluation metrics. Gemini is the leading closed-source model across most tasks, except for the ranking task. Among open-source models, Qwen performs best, ranking second overall and matching or exceeding Gemini on several tasks, including ranking, suggesting that current closed-source models may underperform on ranking tasks.

\begin{table}[!htb]
\centering
\setlength{\tabcolsep}{2pt}
\begin{tabular}{@{}lccccccc@{}}
\toprule
\textbf{} &
  \textbf{Bool} &
  \textbf{SE} &
  \textbf{Count} &
     \multicolumn{2}{c}{\textbf{List}} & 
     
  \textbf{Rank} &
  \textbf{Rsn} \\

\textit{Metrics} &\textit{Acc}$\uparrow$ & \textit{EM} $\uparrow$& \textit{EM} $\uparrow$ & \textit{Prec}$\uparrow$ & \textit{Rec} $\uparrow$ & \textit{RWP} $\uparrow$ & \textit{EM} $\uparrow$ \\    

\midrule
\multicolumn{8}{c}{\textbf{Open Source}} \\ 
\midrule
\textbf{Aya}       & 67.3 & 16.5 & 16.8 & 16.1 & 13.4 & 10.5 & 19.7 \\
\textbf{Deepseek}  & 67.5 & 22.9 & 20.5 & 23.7 & 19.0 & 34.1 & 15.5 \\
\textbf{Intern}    & 69.0 & 22.4 & 21.1 & 22.7 & 19.5 & 47.0 & 25.3 \\
\textbf{Idefics}   & 67.2 & 22.4 & 19.7 & 22.3 & 17.2 & 36.0 & 21.6 \\
\textbf{Llama}     & 69.1 & 22.7 & 19.2 & 25.7 & 22.1 & 21.9 & 24.3 \\
\textbf{Mistral}   & 67.5 & 23.6 & 19.8 & 20.0 & 16.6 & 47.6 & 23.9 \\
\textbf{Molmo}     & 65.0 & 19.1 & \cellcolor[HTML]{CBCEFB}\bf 25.2 & 19.7 & 13.7 & 46.3 & 20.5 \\
\textbf{Qwen}      & \cellcolor[HTML]{CBCEFB}\bf 72.2 & \cellcolor[HTML]{CBCEFB}\bf 26.9  & 23.0 & \cellcolor[HTML]{CBCEFB}\bf 30.7 & 
                     \cellcolor[HTML]{CBCEFB}\bf 21.5 & \cellcolor[HTML]{CBCEFB}\bf 49.5 & 
                     \cellcolor[HTML]{CBCEFB}\bf 32.5 \\ 
\midrule
\multicolumn{8}{c}{\textbf{Closed Source}} \\ 
\midrule
\textbf{Gemini 2.5}    & \cellcolor[HTML]{9AFF99}\bf 80.0 & \cellcolor[HTML]{9AFF99}\bf 36.2 & 
                     \cellcolor[HTML]{9AFF99}\bf 37.6 & \cellcolor[HTML]{9AFF99}\bf 37.8 & 
                     \cellcolor[HTML]{9AFF99}\bf 37.6 & 
                     \cellcolor[HTML]{9AFF99}\bf 57.9 & 
                     \cellcolor[HTML]{9AFF99}\bf 49.0 \\
\textbf{GPT 4o} & 75.8 & 32.7 & 27.5 & 36.1 & 30.9 & 57.2 & 31.7 \\ 

\bottomrule
\end{tabular}%
\caption{\textbf{VLM Results for Different Question Types}. Performance of open-source and closed-source models across different question types. 
Highlighted cells indicate the best close-source (green) and best-open source (blue) performance with performance values marked in \textbf{Bold}. \textbf{SE} stands for single entity question types, and \textbf{Rsn} stands for reasoning question types.}
\label{tab:answer-type}
\end{table}

When analyzing performance by task type, a clear progression in difficulty emerges. Boolean questions achieve the highest scores, with accuracy ranging from 65 to 80\% across models, likely due to their binary answer space. Performance drops sharply on counting questions, reflecting known limitations of current models in arithmetic reasoning. Single-entity prediction tasks show moderate success, with scores between 16 and 36\%, where open-source models generally lag behind proprietary ones. Performance declines further on more complex tasks such as list prediction, ranking, and reasoning, which require multi-step inference and fine-grained spatial understanding. These results highlight the widening gap between simple categorical questions and those demanding deeper geospatial reasoning, underscoring the challenges that current vision-language models face in map-based inference. These results indicate that while current VLMs perform well on some simpler tasks, tasks requiring complex visual reasoning over map images remain challenging.

\vspace{-1em}
\paragraph{Dissecting Model Performance Across Map Categories.} To assess whether VLMs perform consistently across different map categories, we present a category-wise analysis for Gemini 2.5 and Qwen 2.5, the top-performing models in Table \ref{tab:map-type-comparison}. Performance varies by map type and answer format: Gemini 2.5 performs best on three of six tasks for marker maps, and on one task each for conflict, isopelth, and multiple maps, despite these categories being smaller in number than network, divisional, and choropleth maps. Surprisingly, performance on choropleth maps is lower, even though previous benchmarks (Table \ref{tab:mapqa_comparison}) largely consist of these maps and may have been used in pretraining closed-source models. This highlights that our dataset presents novel and significant challenges for current VLMs. Qwen 2.5 shows a similar distribution, excelling on reasoning tasks for marker maps rather than choropleth maps.

\begin{table*}[htb]
\centering
\setlength{\tabcolsep}{4pt}
\renewcommand{\arraystretch}{1}
\begin{tabular}{@{}lcccccccc@{}}
\toprule
\textbf{Map Type} & \textbf{Bool} & \textbf{SE} & \textbf{Count} & \multicolumn{2}{c}{\textbf{List}} & \textbf{Rank} & \textbf{Reasoning} \\
  & \textit{Acc} & \textit{EM} & \textit{EM} & \textit{Prec} & \textit{Rec} & \textit{RWP} & \textit{EM} \\
\midrule
Cartogram   & 82.4 / 65.7 & 35.0 / 23.8 & 40.9 / 21.7 & 37.1 / 41.4 & 33.0 / 33.0 & 74.1 / 68.7 & -- / -- \\
Choropleth  & 81.8 / 76.0 & 31.7 / 27.6 & 35.1 / 25.3 & 32.2 / 25.6 & 38.4 / 19.4 & 32.4 / 29.9 & 46.2 / 30.8 \\
Conflict    & 75.7 / 76.6 &  45.5 / 36.4 & 40.3 / 23.4 & \cellcolor[HTML]{9AFF99}\bf 48.9 / 43.3 & \cellcolor[HTML]{9AFF99}\bf 48.6 / 28.9 & 67.9 / 66.2 & 62.5 / 0.0 \\
Divisional  & 76.2 / 68.9 & 38.5 / 25.0 & 32.4 / 21.2 & 41.6 / 30.1 & 40.7 / 20.3 & 62.9 / 50.3 & 40.0 / 20.0 \\
Isopleth    & 83.9 / 82.0 & 27.4 / 19.4 & 32.4 / 22.7 & 41.1 / 39.8 & 46.0 / 26.2 & 50.0 / 50.0 & \makecell{\cellcolor[HTML]{9AFF99} \bf 75.0} / 25.0 \\
Layout      & 79.4 / 74.2 & 34.6 / 24.8 & 45.1 / 27.1 & 26.5 / 22.8 & 29.9 / 19.6 & 54.8 / 44.8 & 58.6 / 37.9 \\
Marker      & \cellcolor[HTML]{9AFF99}\bf 95.7 / 84.0 & \cellcolor[HTML]{9AFF99}\bf 46.5 / 37.2 & 40.1 / 25.6 & 42.1 / 32.6 & 36.7 / 23.4 & \cellcolor[HTML]{9AFF99}\bf 85.5 / 75.3 & 56.8 / \makecell{\cellcolor[HTML]{9AFF99} \bf 40.7} \\
Mixed       & 73.0 / 69.1 & 37.1 / 26.8 & 38.9 / 24.8 & 34.8 / 17.5 & 33.0 / 12.6 & 52.4 / 35.2 & 50.0 / 27.3 \\
Multiple    & 81.9 / 76.2 & 36.8 / 25.7 & \cellcolor[HTML]{9AFF99}\bf 44.6 / 33.9 & 26.5 / 31.8 & 20.2 / 13.8 & 52.8 / 31.1 & 52.9 / 17.6 \\
Network     & 78.5 / 63.9 & 37.9 / 29.2 & 36.4 / 17.8 & 41.1 / 35.8 & 38.1 / 25.4 & 61.7 / 54.9 & 32.4 / 18.9 \\
\bottomrule
\end{tabular}
\caption{\textbf{Performance across different map categories.} Each cell reports \textit{Gemini 2.5 / Qwen 2.5}. Metrics: Accuracy (Bool), Exact Match (SE, Count, Reasoning), Precision / Recall (List), and Mean Rank-wise Precision (Rank) ( ref eq \ref{eq:rank_wise_precision} ).}
\label{tab:map-type-comparison}
\end{table*}

\vspace{-1em}
\paragraph{VLM Performance Across Geographic Scales.}
Table~\ref{tab:geo-level-comparison} shows Gemini 2.5 (G) and Qwen 2.5 VL (Q) performance across twelve hierarchical geographical levels, from building- to world-level maps. Both models perform better on broader levels (Continent, Country) and decline at intermediate and localized levels (County, State, District, Building), where fine-grained distinctions are greater. Visual tasks remain strong, but reasoning-intensive tasks like Ranking and Comparative questions drop noticeably at localized levels. Gemini 2.5 generally outperforms Qwen 2.5 on broader levels, while Qwen 2.5 shows gains on select intermediate levels. Overall, VLMs handle large-scale spatial reasoning more reliably than precise, localized reasoning, highlighting the challenges of detailed map understanding.

\begin{table*}[htb]
\centering
\setlength{\tabcolsep}{4pt}
\renewcommand{\arraystretch}{1}
\begin{tabular}{@{}lcccccccc@{}}
\toprule
\textbf{Geographical} & \textbf{Bool} & \textbf{SE} & \textbf{Count} & \multicolumn{2}{c}{\textbf{List}} & \textbf{Rank} & \textbf{Reasoning} \\
 \textbf{Granularity} & \textit{Acc} & \textit{EM} & \textit{EM} & \textit{Prec} & \textit{Rec} & \textit{RWP} & \textit{EM} \\
\midrule
Building     & 79.1 / 73.6 & 31.6 / 22.4 & 43.1 / 26.9 & 25.6 / 19.5 & 28.9 / 17.3 & 58.1 / 51.3 & 61.5 / 23.1 \\
Campus       & 80.0 / 76.4 & 38.1 / 27.7 & \makecell{\cellcolor[HTML]{9AFF99}\textbf{48.3} }/ 26.3 & 27.4 / 28.2 & 30.6 / 23.3 & 49.8 / 35.3 & 56.3 / \makecell{\cellcolor[HTML]{9AFF99}\textbf{50.0}} \\
Neighborhood & 81.3 / 54.7 & \cellcolor[HTML]{9AFF99}\bf 49.5 / 37.6 & 37.0 / 18.5 & 48.5 / 30.5 & 45.7 / 21.7 & 51.6 / 53.4 & 20.0 / 20.0 \\
District     & 71.4 / 71.4 & 24.5 / 15.4 & 41.3 / 23.9 & 43.0 / 34.2 & 37.8 / 27.0 & 10.3 / 6.4  & 50.0 / 20.0 \\
City         & 75.7 / 69.4 & 36.7 / 25.5 & 34.4 / 18.1 & 32.9 / 23.3 & 30.4 / 15.8 & 70.3 / 53.4 & 41.9 / 27.4 \\
County       & 80.0 / 45.0 & 34.6 / 26.3 & 34.6 / \makecell{\cellcolor[HTML]{9AFF99}\bf{42.3}} & \makecell{\cellcolor[HTML]{9AFF99}\textbf{55.7}} / 11.1 & \makecell{\cellcolor[HTML]{9AFF99}\textbf{48.2}} / 5.6  & 36.2 / 36.1 & -- / -- \\
State        & 81.7 / 73.3 & 44.1 / 30.4 & 32.0 / 24.0 & 23.2 / 10.6 & 25.8 / 7.1  & \cellcolor[HTML]{9AFF99}\bf 80.6 / 68.1 & -- / -- \\
Region       & 80.6 / 75.8 & 34.4 / 22.8 & 37.5 / 22.9 & 40.1 / 37.6 & 43.4 / 27.8 & 68.6 / 61.2 & 55.6 / 16.7 \\
Country      & 81.9 / 73.6 & 35.5 / 27.4 & 38.3 / 24.7 & 41.6 / 30.7 & 41.5 / 21.7 & 57.2 / 47.7 & 54.6 / 35.4 \\
Subcontinent & 73.4 / 74.7 & 39.9 / 31.1 & 35.3 / 23.5 & 45.9 / \makecell{\cellcolor[HTML]{9AFF99}\textbf{44.7}} & 47.5 / \textbf{32.7} & 60.2 / 48.0 & 50.0 / 0.0 \\
Continent    & \cellcolor[HTML]{9AFF99}\bf 86.5 / 77.5 & 39.7 / 30.6 & 32.8 / 22.7 & 42.4 / 41.3 & 37.6 / 22.7 & 63.6 / 48.8 & \makecell{\cellcolor[HTML]{9AFF99}\textbf{65.4}} / 34.6 \\
World        & 79.9 / 71.5 & 35.2 / 28.3 & 37.8 / 22.4 & 36.5 / 32.0 & 38.0 / 21.8 & 48.9 / 48.2 & 44.2 / 30.2 \\
\bottomrule
\end{tabular}
\caption{\textbf{Performance across different geographical granularities} Each cell reports \textit{Gemini 2.5 / Qwen 2.5}. Metrics: Accuracy (Bool), Exact Match (SE, Count, Reasoning), Precision / Recall (List), and Mean Rank-wise Precision (Rank) ( ref eq \ref{eq:rank_wise_precision} ).}
\label{tab:geo-level-comparison}
\end{table*}

\vspace{-1em}
\paragraph{Performance Across Reasoning Classes.} We evaluate Gemini 2.5, the best-performing closed-source model, across different reasoning classes, each posing distinct challenges (Table \ref{tab:question-type-combined}). Visual questions achieve the highest accuracy across most tasks, except for Ranking, indicating strong handling of perceptual features such as color, pattern, and iconography. Spatial questions follow, while Comparative questions show the lowest performance, reflecting difficulties with multi-step reasoning and sub-tasks like Single Entity and Counting. Questions combining multiple aspects (Visual + Comparative + Spatial) perform the worst, with metrics near zero, highlighting the challenges of multi-faceted reasoning. In contrast, questions labeled as Other—typically reliant on metadata or contextual cues—achieve intermediate performance, suggesting that the model leverages auxiliary information from large-scale pre-training rather than performing true complex reasoning.

\begin{table*}[!htb]
\centering
\setlength{\tabcolsep}{4pt}
\begin{tabular}{@{}lccccccc@{}}
\toprule
\textbf{} &
  \multicolumn{1}{c}{\textbf{Bool}} &
  \multicolumn{1}{c}{\textbf{SE}} &
  \multicolumn{1}{c}{\textbf{Count}} &
  \multicolumn{2}{c}{\textbf{List}} &
  \multicolumn{1}{c}{\textbf{Rank}} &
  \multicolumn{1}{c}{\textbf{Rsn}} \\
\textit{Class} & \textit{Acc} & \textit{EM} & \textit{EM} & \textit{Prec} & \textit{Rec} &\textit{RWP}  & \textit{EM} \\
\midrule
\textbf{Visual}       & 80.7 / 76.0 & 32.7 / 28.3 & 36.9 / 20.3 & 39.2 / 22.8 & 38.7 / 21.5 & 0.0 / 79.6 & 64.5 / 29.0 \\
\textbf{Comparative}  & 80.5 / 66.5 & \cellcolor[HTML]{9AFF99}\bf 30.0 / 22.4 & 23.8 / 24.2 & 39.1 / \makecell{\cellcolor[HTML]{9AFF99}\bf 35.9} & \makecell{\cellcolor[HTML]{9AFF99}\bf 41.6} / 24.8 & 16.3 / 68.6 & 55.6 / 22.2 \\
\textbf{Spatial}      & 76.9 / 68.0 & 38.0 / 25.3 & 31.7 / 16.3 & 39.5 / 31.1 & 38.3 / 20.4 & 13.3 / 90.2 & 40.6 / 66.7 \\
\textbf{V+C}          & 81.2 / 77.8 & 40.0 / 28.3 & 38.8 / 17.9 & 37.8 / 33.2 & 36.4 / 23.3 & \cellcolor[HTML]{9AFF99}\bf 56.2 / 80.9 & \cellcolor[HTML]{9AFF99}\bf 66.7 / 34.8 \\
\textbf{V+S}          & 80.7 / 74.0 & 35.4 / 22.9 & 33.9 / 18.2 & 40.4 / 32.5 & 39.9 / 23.1 & 0.0 / 79.4 & 23.1 / 7.7 \\
\textbf{C+S}          & \cellcolor[HTML]{9AFF99}\bf 83.7 / 74.1 & 35.4 / 26.7 & \cellcolor[HTML]{9AFF99}\bf 52.0 / 33.3 & \makecell{\cellcolor[HTML]{9AFF99}\bf 40.5} / 27.0 & 39.5 / 18.7 & 19.9 / 74.4 & 36.4 / 22.2 \\
\textbf{V+C+S}        & 81.3 / 79.2 & 39.8 / 28.4 & 38.1 / 16.3 & 32.4 / 30.0 & 28.5 / 20.2 & 0.0 / 81.4 & 14.3 / 0.0 \\
\textbf{Other}        & 79.9 / 71.3 & \cellcolor[HTML]{9AFF99}\bf 38.4 / 32.0 & 40.8 / 22.1 & 39.4 / 34.3 & 39.7 / \makecell{ \cellcolor[HTML]{9AFF99}\bf 25.0} & 15.2 / 74.2 & 52.3 / 36.4 \\
\bottomrule
\end{tabular}
\caption{\textbf{Performance across different reasoning classes.} Each cell reports \textit{Gemini 2.5 / Qwen 2.5}. Metrics: Accuracy (Bool), Exact Match (SE, Count, Reasoning), Precision / Recall (List), and Mean Rank-wise Precision (Rank).  V = visual, C = comparative, S = spatial.}
\label{tab:question-type-combined}
\end{table*}

\vspace{-1em}
\paragraph{Low Text-Only Performance Indicates Limited Memorization of \datasetName.} Table \ref{tab:open-vs-closed_text_only} presents model performance in text-only mode, revealing a substantial drop across nearly all question types. Boolean questions are a notable exception, achieving roughly 60\% accuracy even without visual input, suggesting that these questions may be inherently easier or partially addressed by knowledge acquired during pretraining. In contrast, tasks requiring richer reasoning—such as Single Entity identification, Counting, Listing, Ranking, and general Reasoning—show dramatic performance declines, indicating that visual cues are essential for solving these questions. This disparity highlights the robustness of \datasetName ~as a benchmark: it effectively distinguishes between tasks that can be addressed through memorization or prior knowledge and those that require true multi-modal visual-language reasoning. Therefore \datasetName ~provides a rigorous testbed for evaluating VLM capabilities in integrating visual and textual understanding.

\begin{table}[htb]
\centering
\setlength{\tabcolsep}{3pt}
\renewcommand{\arraystretch}{1.0}
\begin{tabular}{@{}lrrrrrrr@{}}
\toprule
\textbf{Model} & \textbf{Bool} & \textbf{SE} & \textbf{Count} & \multicolumn{2}{c}{\textbf{List}} & \textbf{Rank} & \textbf{Rsn} \\
 & \textit{Acc} & \textit{EM} & \textit{EM} & Prec. & Rec. & RWP & EM \\
\midrule
\multicolumn{8}{c}{\textbf{Open Source}} \\
\midrule
Aya       & { \cellcolor[HTML]{CBCEFB}\bf 63.0} & { \cellcolor[HTML]{CBCEFB}\bf 6.3} & { \cellcolor[HTML]{CBCEFB}\bf 5.9} & 4.6 & 3.4 & 3.2  & 6.3 \\
Deepseek  & 61.4 & 3.6 & 0.2 & 1.1 & 1.1 & 22.8 & 6.4 \\
Intern    & 61.5 & 2.1 & 1.4 & 1.0 & 1.1 & 14.6 & 7.09 \\
Idefics   & 61.8 & 4.1 & 3.4 & 1.6 & 2.9 & 19.4 & 10.6 \\
Llama     & 62.3 & 2.1 & 2.0 & 1.7 & 1.4 & 4.9  & 9.1 \\
Molmo     & 61.4 & 7.0 & 1.5 & { \cellcolor[HTML]{CBCEFB}\bf 8.6} & { \cellcolor[HTML]{CBCEFB}\bf 5.9} & { \cellcolor[HTML]{CBCEFB}\bf 38.8} & { \cellcolor[HTML]{CBCEFB}\bf 15.2} \\
Mistral   & 61.2 & 0.3 & 0.3 & 0.1 & 0.1 & 1.9  & 7.4 \\
Qwen      & 61.2 & 5.1 & 2.0 & 3.5 & 2.1 & 32.0 & 13.1 \\
\midrule
\multicolumn{8}{c}{\textbf{Closed Source}} \\
\midrule
Gemini    & \cellcolor[HTML]{9AFF99}\bf 62.9 & 3.8 & \cellcolor[HTML]{9AFF99}\bf 5.2 & \cellcolor[HTML]{9AFF99}\bf 3.9 & \cellcolor[HTML]{9AFF99}\bf 5.0 & 21.7 & 14.1 \\
GPT-4o    & 62.2 & \cellcolor[HTML]{9AFF99}\bf 6.0 & 2.4 &  3.1 & 2.4 & \cellcolor[HTML]{9AFF99}\bf 44.4 & \cellcolor[HTML]{9AFF99}\bf 20.27 \\
\bottomrule
\end{tabular}
\caption{\textbf{Performance across Open-Source and Closed-Source Models in Text only Mode.} Accuracy (Bool), Exact Match (SE, Count, Reasoning), Precision / Recall (List), and Rank-wise Precision (Rank). Dashes (--) indicate N/A. \textbf{SE} stands for single entity question types, and \textbf{Rsn} stands for reasoning question types.}
\label{tab:open-vs-closed_text_only}
\end{table}

\vspace{-1em}
\paragraph{Ablation Experiments.} This section examines the robustness of VLMs to input data corruption and its effect on model inference. We focus on the best-performing open-source model, Qwen-2.5-VL-7B-Instruct, a type-agnostic, easily deployable model that supports efficient inference and operates on the image's native resolution, making it well-suited for systematically assessing the impact of data variations on performance.

\vspace{-1em}
\paragraph{Pertubation Ablation.}
Focusing on real-world, often noisy data, we investigate the impact of noise injection on model performance, hypothesizing that pixel-level perturbations affect the visual encoder’s ability to extract information. We evaluate three noise types: (1) random Gaussian noise with magnitude $[-10\%, 10\%]$ of the pixel range, (2) random pepper noise affecting 10\% of pixels, and (3) a black rectangle masking 10\% of the image area at a random location, with aspect ratios sampled between 0.5 and 2.0 to avoid extreme shapes. Table \ref{tab:qwen_perturb_compact} shows that Gaussian noise has minimal effect, likely due to pretrained models using similar augmentation. Pepper noise impacts performance more than localized black rectangles, suggesting distributed information loss is more detrimental than localized occlusion—contrasting human visual behavior. Performance drops vary by question type: reasoning-heavy questions (Single Entity, Counting, Reasoning, List) decline significantly, while Boolean and Ranking questions remain largely unaffected.

\vspace{-1em}
\paragraph{Resolution Ablation.}
To examine the effect of input quality on model performance, we systematically degrade image resolution. The model is evaluated at the original resolution, followed by its resized versions having 50\% reduction and a further 25\% reduction in image dimensions. We employed the Lanczos resampling technique to produce sharp images with minimal aliasing while preserving fine details. VLM performance declines with lower image quality, with drops of up to 50\% in relative score. Tasks that rely on detailed visual information—such as Single Entity, List, Counting, and Reasoning—show the most pronounced degradation, highlighting the model’s sensitivity to fine-grained visual cues for these categories.


\section{Conclusion and Future Work}
We introduce \datasetName, a large-scale benchmark of real-world maps with human-authored questions, designed to evaluate VLMs’ geospatial reasoning capabilities.  Accuracy across different answer formats remains low even for state-of-the-art models, underscoring the significant challenges that VLMs face and highlighting the substantial potential for improvement. This reinforces the value of \datasetName ~as a benchmark dataset for evaluating and advancing multimodal reasoning capabilities. Our group-wise analyses across map categories, geographic granularity, and reasoning classes show that while current VLMs handle general visual question answering competently, they struggle with advanced geospatial reasoning requiring a deeper understanding of spatial relationships and context. We show that VLMs are robust to Gaussian noise but sensitive to pepper noise and random rectangular occlusions, with the greatest drops observed under random pepper noise. These findings expose limitations in existing models and point to promising research directions. A key limitation of our study is reliance on manually curated prompts and fixed few-shot examples, which are labor-intensive and less scalable. Expanding \datasetName ~to cover additional map categories and broader visual reasoning tasks, including temporal data, alongside exploring novel modeling and prompting strategies or fine-tuning on this dataset, offers avenues to improve geospatial reasoning in VLMs. 


\newpage
{
    \small
    \bibliographystyle{ieeenat_fullname}
    \bibliography{main}

}
\newpage
\clearpage
\setcounter{figure}{0}
\renewcommand{\thefigure}{S\arabic{figure}}
\setcounter{section}{0}
\renewcommand{\thesection}{S\arabic{section}}
\setcounter{table}{0}
\renewcommand{\thetable}{S\arabic{table}}
\section{Supplementary}

\subsection{Additional Details on \datasetName ~Collection and Annotation} \label{ssec:web_searches}
To ensure diversity and realism, we curated maps from publicly available sources across the internet, including news portals, educational resources, weather forecast platforms, and other open forums. For each map type, targeted Google searches were performed using specific keywords to retrieve a large initial pool of candidate maps. The search queries included terms such as:

\begin{itemize}
    \item \textbf{Network Maps:} , ``Ferry route maps'', ``School bus routes map'',  ``Human migration route maps'', ``MigrantMaps'' , “High quality metro map”
    \item \textbf{Divisional Maps:} ``Divisional maps for pincodes Africa'', ``Divisional maps for pincodes Australia'', ``Divisional maps for pincodes Europe'', ``DivisionMapsSchoolDistrict'' , ``Time zones map'' 
    \item \textbf{Choropleth Maps:} ``Choropleth'', ``choropleth map USA'', ``choropleth map world'', ``choropleth maps others'', ``multivariate choropleth maps''  
    \item \textbf{Layout Maps:} ``Memorial hospital layout maps'', ``Stadium layout maps''  
    \item \textbf{Marker Maps:} ``Top hospitals in California marker maps'',``Dot\_Maps'', ``World\_Heritage'',   
    \item \textbf{Mixed Maps:} ``Printable city tourist attractions map'', ``Sightseeing map'',   
    \item \textbf{Isopleth Maps:} ``Contour maps'', ``Isopleth weather maps'', ``Isopleth'' ,  ``climate'', ``WeatherMapsIso''
    \item \textbf{Multiple Maps:} ``Bivariate map'', ``ComparisonMaps''  
    \item \textbf{Conflict Maps:} ``Territory changes in conflict map'' 
    \item \textbf{Cartograms:} ``Cartogram maps'', ``cartogram in statistics''  
\end{itemize}

While scraping the data from the internet, we browsed through all the maps images and then the final set of maps were selected based on human readability and resolution. Also note that while these queries were made with the intention of retrieving those candidate maps of the search query categories, we have manually verified removed noisy, low resolution, blurry, cropped or irrelevant images in the process along with reclassifying the misclassified ones. 

\paragraph{Additional Annotation Details.} Maps could be independently annotated by multiple annotators, each generating their own questions and answers without access to others’ work. This intentional redundancy was designed to capture variability in human interpretation and improve QA quality. Even when the same map was assigned to multiple annotators, each produced a distinct set of questions; consequently, we did not encounter cases where annotators generated identical questions. Therefore, the issue of disagreements in answers does not apply here.

\subsection{Additional \datasetName ~taxonomy} \label{ssec:addtional_taxonomy}
Table \ref{tab:image_size_distribution} provides a detailed breakdown of the number of images in \datasetName ~across various resolutions, highlighting the prevalence of each size. Table \ref{tab:geographic_level_distribution} further characterizes the dataset by showing how images are distributed according to different levels of geographic granularity, offering insight into the spatial coverage and diversity of the dataset.

\begin{table}[!htb]
    \centering
    \begin{tabular}{lr}
        \toprule
        \textbf{Image Size} & \textbf{Count} \\
        \midrule
        $>$8MP & 79 \\
        2--8MP & 212 \\
        0.5--2MP & 465 \\
        0.1--0.5MP & 250 \\
        $<$0.1MP & 19 \\
        \bottomrule
    \end{tabular}
    \caption{\textbf{Image Size Distribution by Megapixels (MP) in \datasetName}.}
    \label{tab:image_size_distribution}
\end{table}

\begin{table}[!htb]
\small
    \centering
    \begin{tabular}{lr}
        \toprule
        \textbf{Geographic Level} & \textbf{Count} \\
        \midrule
        World        & 168 \\
        Continent    & 68 \\
        Subcontinent & 49 \\
        Country      & 264 \\
        State        & 32 \\
        Region       & 121 \\
        County       & 14 \\
        City         & 147 \\
        District     & 28 \\
        Neighborhood & 30 \\
        Campus       & 42 \\
        Building     & 62 \\
        \bottomrule
    \end{tabular}
    \caption{\textbf{Distribution of Maps Across Different Geographic Granularity}.}
    \label{tab:geographic_level_distribution}
\end{table}

\begin{figure}[!htb]  
    \centering
    \includegraphics[width=0.45\textwidth]{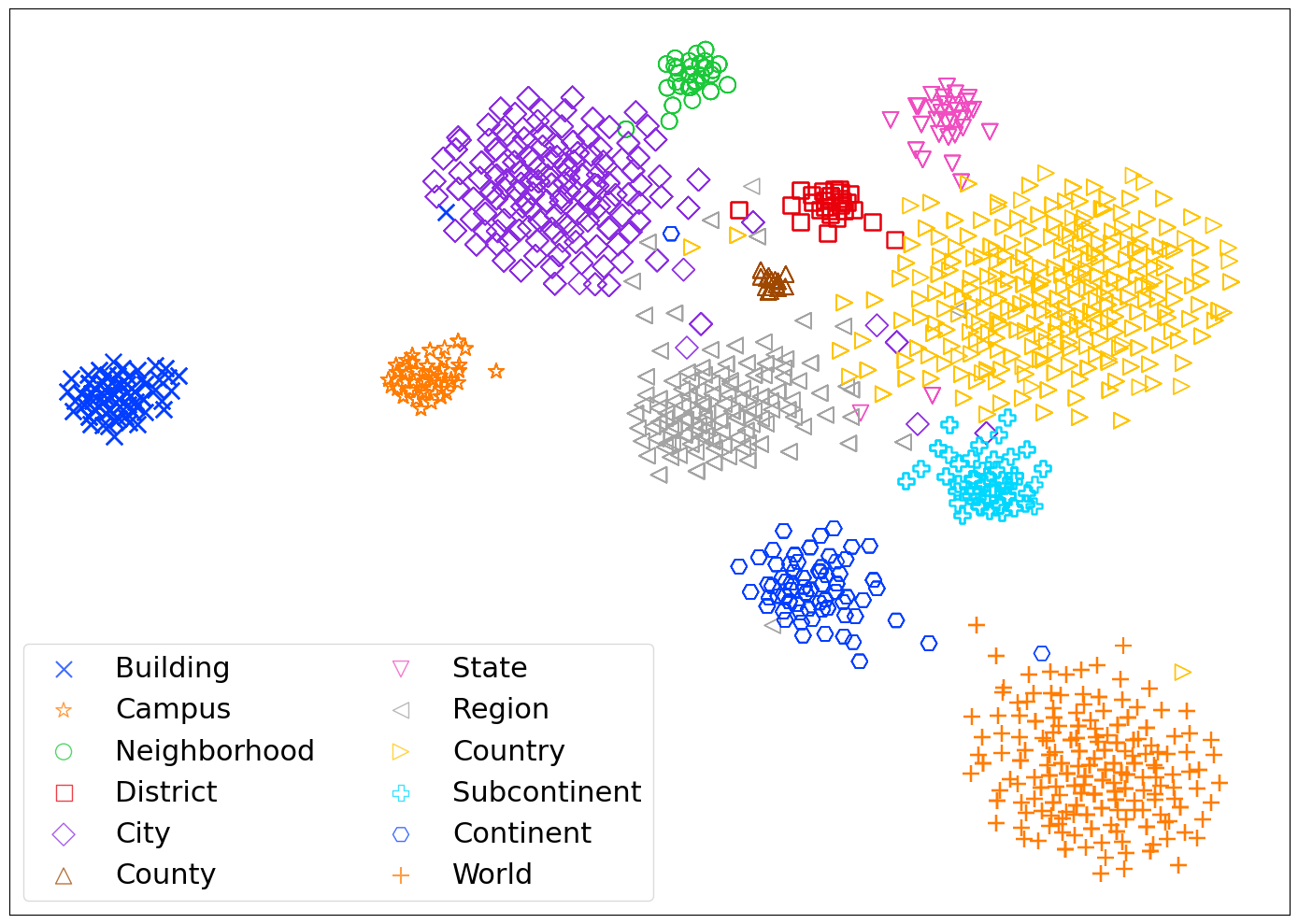} 
    \caption{\textbf{T-SNE for images geographical granularity.}}
    \label{fig:tsne-geographical-level}
\end{figure}


\begin{figure}[h!]
    \centering
    \begin{subfigure}[b]{\linewidth}
        \centering
        \includegraphics[width=\linewidth]{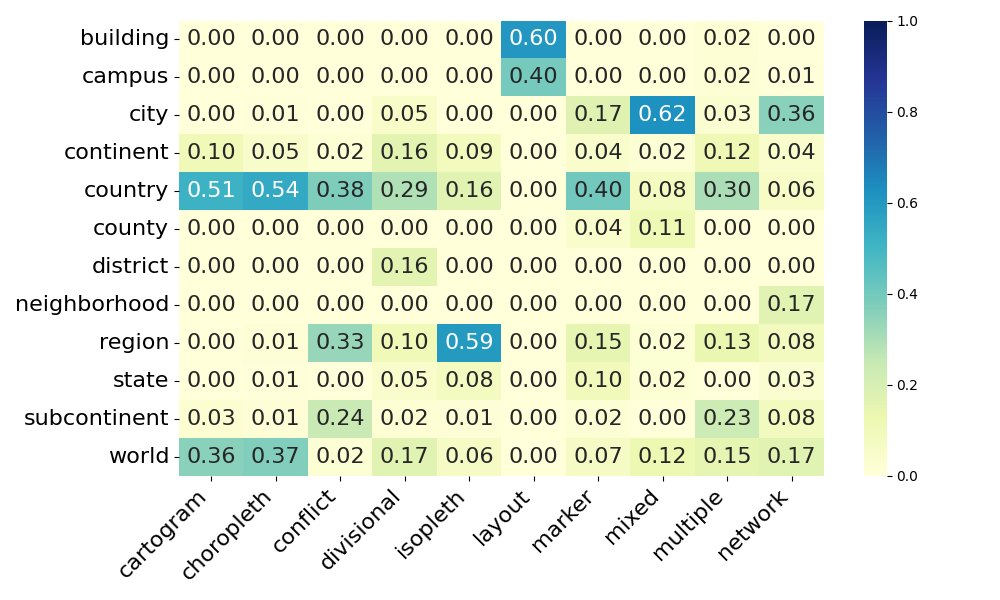}
        \caption{Relative heatmap normalized by map type.}
        \label{fig:sub-map-type}
    \end{subfigure}
    
    \begin{subfigure}[b]{\linewidth}
        \centering
        \includegraphics[width=\linewidth]{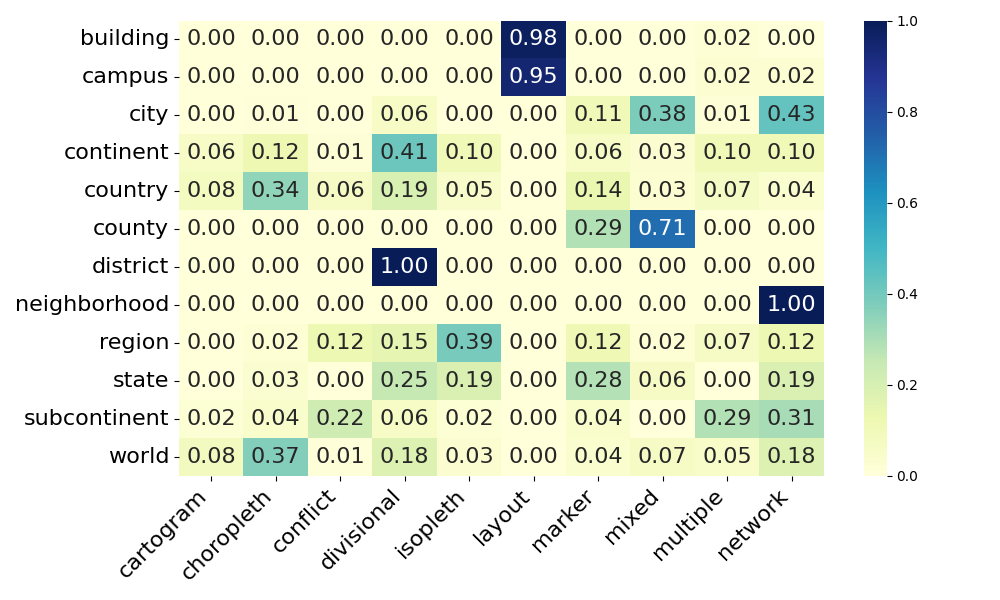}
        \caption{Relative heatmap normalized by geographical granularity.}
        \label{fig:sub-geographic-level}
    \end{subfigure}
    
    \caption{Correlation heatmaps showing the proportional relationship between map type (X-axis) and geographic level (Y-axis)}
    \label{fig:combined-heatmaps}
\end{figure}

Figure \ref{fig:combined-heatmaps} shows two heatmaps that provide a comprehensive analysis of the distribution of geographical granularity across different map types in our dataset. This analysis highlights significant structural biases and specialization within the data.

We normalize the heatmap in Figure \ref{fig:sub-map-type} by map type and show the proportional distribution of geographic levels within each map category. The data reveals a strong trend of specialization. For instance, layout maps are almost exclusively associated with the building and campus levels, while choropleth and isopleth maps have been frequently found at the country level. In contrast, map types like mixed and network show a more diverse spread across various geographic levels, indicating their versatility for representing a wider range of spatial scales.

The heatmap in Figure \ref{fig:sub-map-type}, normalized by geographical granularity, offers a complementary perspective by showing which map types are most prevalent within a specific geographic level. The data highlights a clear dominance of certain map types at particular granularity. For example, layout maps constitute nearly all content at the building and campus levels, confirming their conventional use for these finer granularities. Similarly, divisional and network maps dominate at district and neighborhood levels respectively. However, some geographic granularities, such as country and state, are more diverse, with significant representation from multiple map types. This suggests a richer variety of mapping tasks and conventions are employed at these scales. 

\subsection{Evaluation Metrics}
Rank-Wise Precision (RWP) evaluates a model's ranking ability by considering not only the order of elements but also instances where the model may add or delete elements from the set. This differs from simple rank-based metrics which assume a fixed set of elements and only measure misalignment. It is defined as the average of the $Precision@K$ for each valid value of $K$. The formula is given by:
\begin{equation}
\label{eq:rank_wise_precision}
\text{RWP} = \frac{1}{M} \sum_{k=1}^{M} \text{Precision@k}
\end{equation}
where $L_{pred}$ is the model-predicted ranked list, $|L_{ref}|$ is the correct ranked list and $M = max(|L_{pred}|,|L_{ref}|)$ with $|L|$ denoting length of list $L$.

\subsection{Ablation Results}
Table \ref{tab:qwen_perturb_compact} presents the results of our ablation study examining how various types of image corruption affect the performance of VLMs, highlighting the model’s sensitivity to degraded visual input. 

Similarly, Table \ref{tab:qwen_resolution_compact} illustrates the impact of reduced image resolution on VLM performance, demonstrating how lower-quality visual data can lead to substantial drops in accuracy across different tasks, especially for reasoning questions.  Although the dataset includes map images with varying native resolutions, we did not apply any manual resizing or preprocessing. All images were loaded as PIL objects and passed directly to the corresponding Hugging Face AutoProcessor, which automatically performs any required transformations—such as resizing or padding—to match the vision encoder’s input specifications. Thus, while the models accept images of arbitrary native resolution, all rescaling is handled internally by the models’ official preprocessing pipelines.

\begin{table}[!htb]
\centering
\setlength{\tabcolsep}{3pt}
\begin{tabular}{@{}lcccccc@{}}
\toprule
\textbf{} & \textbf{Bool} & \textbf{SE} & \textbf{Count} & \textbf{List (P/R)} & \textbf{Rank} & \textbf{Rsn} \\
\textit{Metrics} & \textit{EM} & \textit{EM} & \textit{EM} & \textit{Prec / Rec} & \textit{RWP} & \textit{EM} \\
\midrule
Overall            & 72.2 & 26.8 & 23.0 & 30.4 / 21.3 & 49.6 & 30.7 \\
RN       & 71.5 & 26.4 & 22.7 & 29.7 / 21.1 & 50.0 & 29.1 \\
RPN    & 70.9 & 21.6 & 20.9 & 24.5 / 17.0 & 48.5 & 24.0 \\
RBR  & 71.5 & 24.9 & 21.8 & 29.0 / 19.7 & 49.2 & 27.0 \\
\bottomrule
\end{tabular}
\caption{\textbf{QWEN 2.5 VL – Perturbation Ablation.} RN = Random Noise, RPN = Random Pepper Noise, RBR = Random Black Rect,EM = Exact Match, RWP = Rank-wise Precision. List column shows Precision / Recall.}
\label{tab:qwen_perturb_compact}
\vspace{-1.5em}
\end{table}

\begin{table}[!htb]
\centering
\setlength{\tabcolsep}{2pt}
\begin{tabular}{@{}lcccccc@{}}
\toprule
\textbf{} & \textbf{Bool} & \textbf{SE} & \textbf{Count} & \textbf{List (P/R)} & \textbf{Rank} & \textbf{Rsn} \\
\textit{Metrics} & \textit{EM} & \textit{EM} & \textit{EM} & \textit{Prec / Rec} & \textit{ RWP} & \textit{EM} \\
\midrule
Orig. Res & 72.2 & 26.8 & 23.0 & 30.4 / 21.3 & 49.6 & 30.7 \\
50\% drop     & 70.5 & 21.9 & 20.3 & 25.3 / 17.3 & 48.3 & 25.3 \\
75\% drop    & 69.1 & 16.1 & 17.9 & 19.2 / 13.0 & 45.4 & 15.2 \\
\bottomrule
\end{tabular}
\caption{\textbf{QWEN 2.5 VL – Resolution Ablation.} EM = Exact Match, RWP = Rank-wise Precision. List column shows Precision / Recall.}
\label{tab:qwen_resolution_compact}
\end{table}

\subsection{Error Analysis}

Our error analysis shows that model failures in MAPVERSE are driven by a combination of \textbf{misperception}, \textbf{misgrounding}, and \textbf{faulty spatial or comparative reasoning}, with each map category amplifying different aspects of these weaknesses. Across \textbf{choropleth, isopleth, and cartogram maps}, models frequently misinterpret legends, confuse visually similar color gradients, or conflate visual area with encoded values, leading to systematic mistakes in comparisons and trend identification. \textbf{Network, layout, and divisional maps} expose limitations in spatial reasoning: models misread topological structure, misidentify region boundaries or adjacency, and produce incorrect multi-step routes or positional relationships. In \textbf{marker and mixed maps}, errors arise from visual clutter—missed or double-counted markers, confusion between symbol categories, and failure to separate overlapping visual layers. \textbf{Conflict and multi-layer maps} further amplify these issues, with models often blending information across layers or misinterpreting symbolic icons. Finally, \textbf{multiple-map layouts} introduce consistent cross-panel grounding errors, where the model retrieves information from the wrong map or fails to integrate evidence across panels. Taken together, these patterns indicate that VLMs struggle when maps require \textbf{precise legend use, robust visual grounding, or compositional spatial reasoning}, and that error modes recur across categories whenever maps demand multi-layer integration or fine-grained visual discrimination.

\subsection{Additional \datasetName ~ question answers Samples}

\begin{figure}[!htb]
    \centering
    \includegraphics[width=1.0\linewidth]{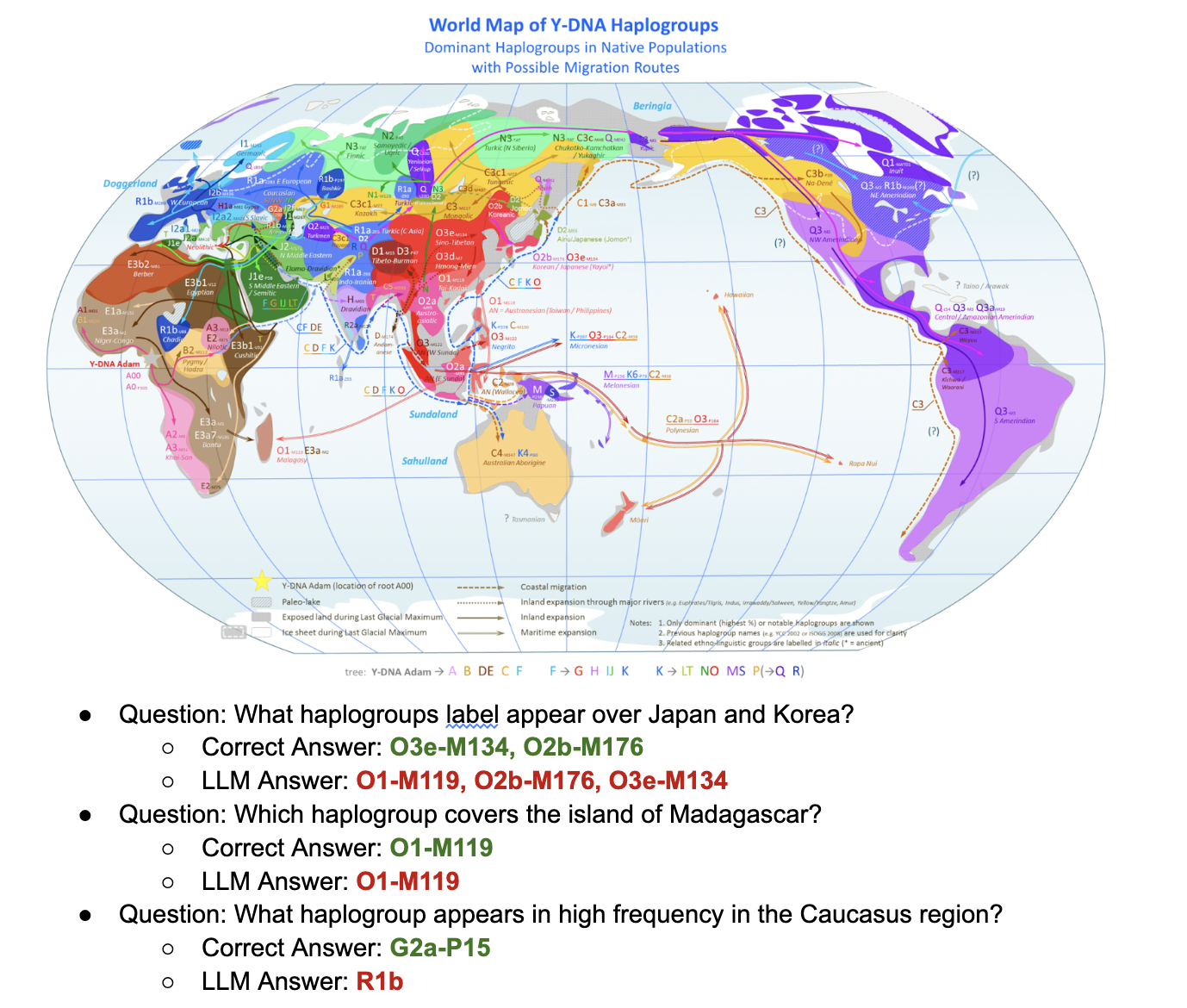}
    \caption{Sample QA for mixed map type (isopleth + network)}
\end{figure}

\begin{figure}[!htb]
    \centering
    \includegraphics[width=1.0\linewidth]{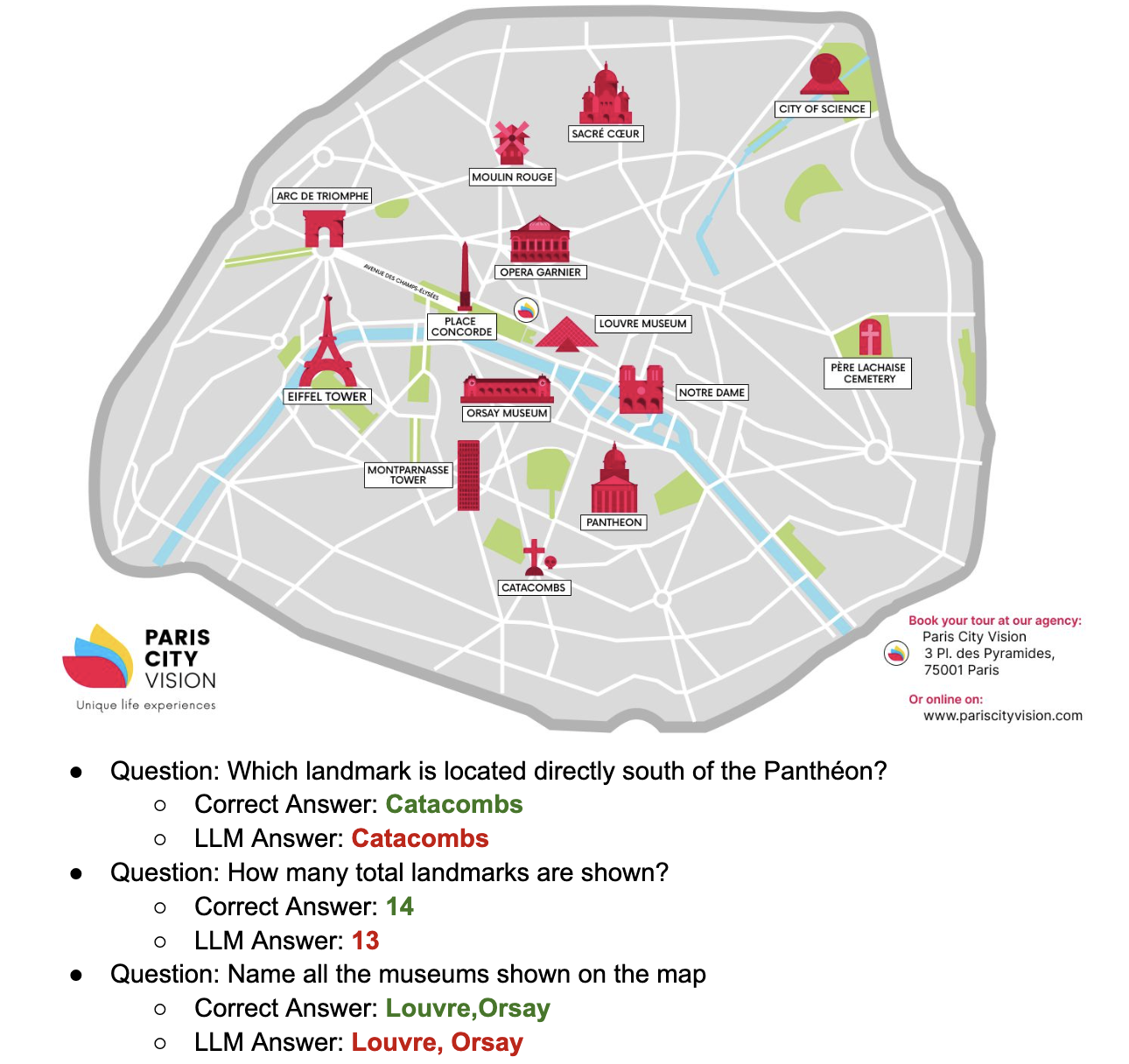}
    \caption{Sample QA for mixed map type (marker + network)}
\end{figure}

\begin{figure}[!htb]
    \centering
    \includegraphics[width=1.0\linewidth]{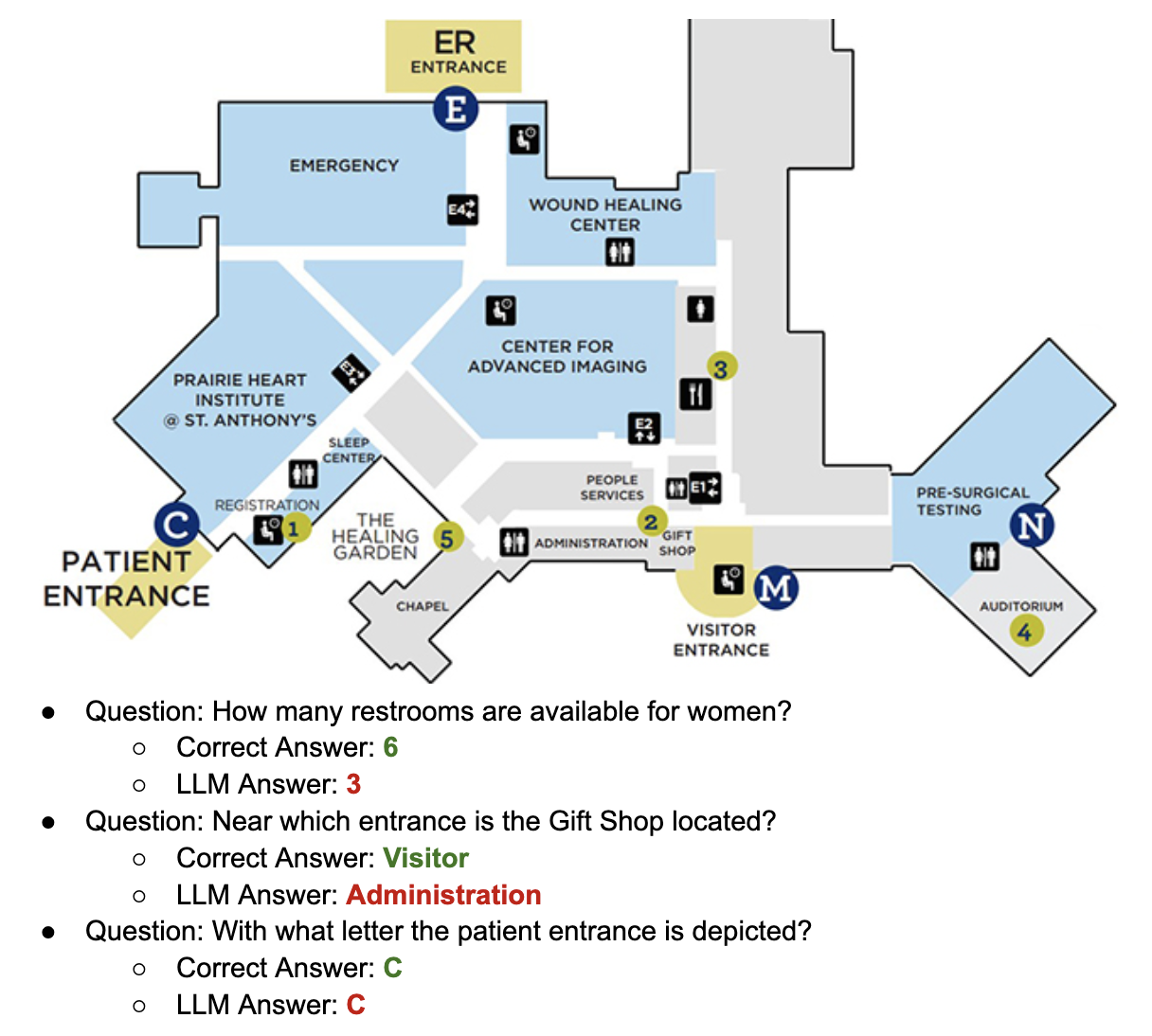}
    \caption{Sample QA for layout map type }
\end{figure}

\begin{figure}[!htb]
    \centering
    \includegraphics[width=1.0\linewidth]{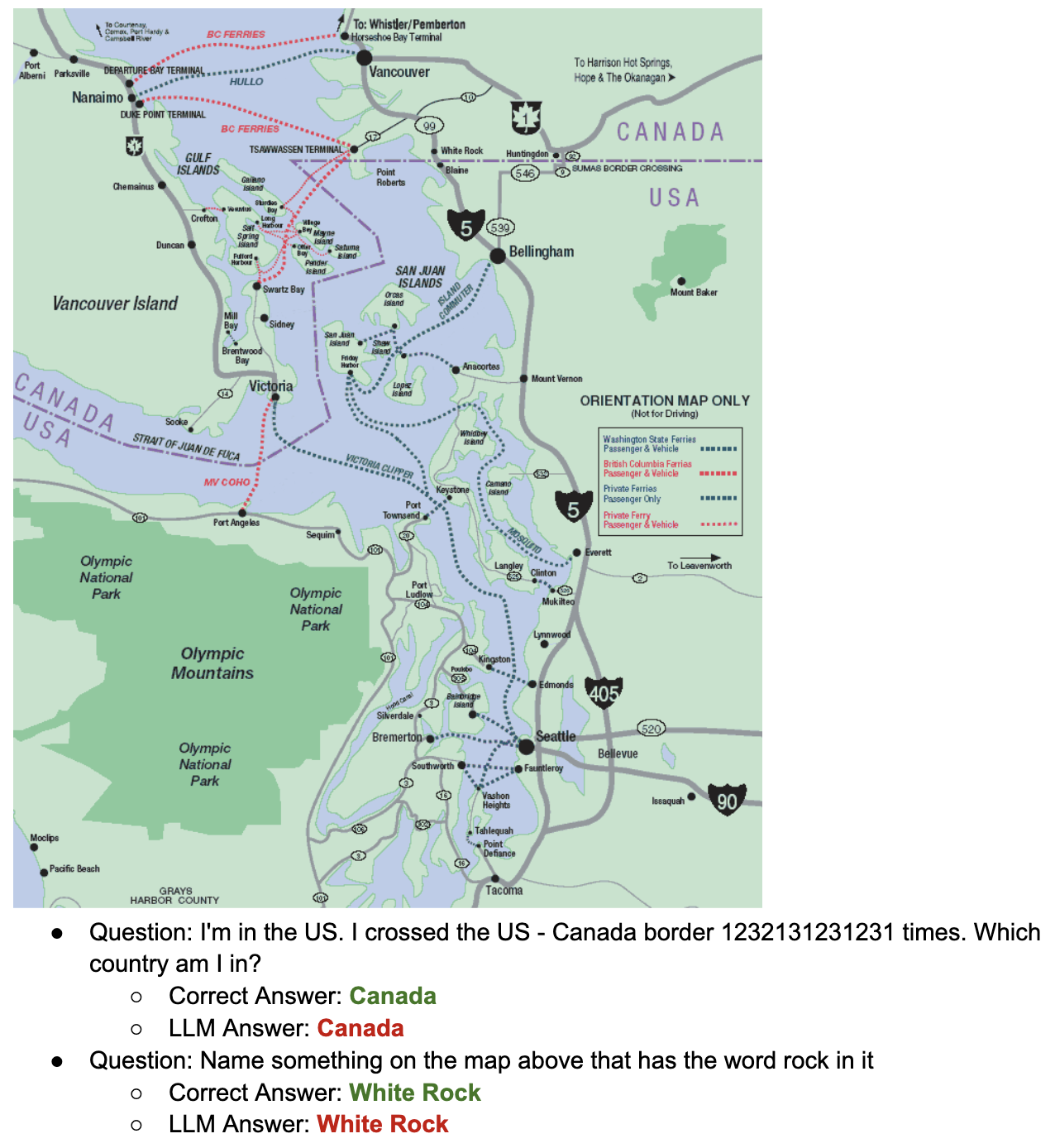}
    \caption{Sample QA for mixed map type (isopleth + network + marker)}
\end{figure}

\begin{figure}[!htb]
    \centering
    \includegraphics[width=1.0\linewidth]{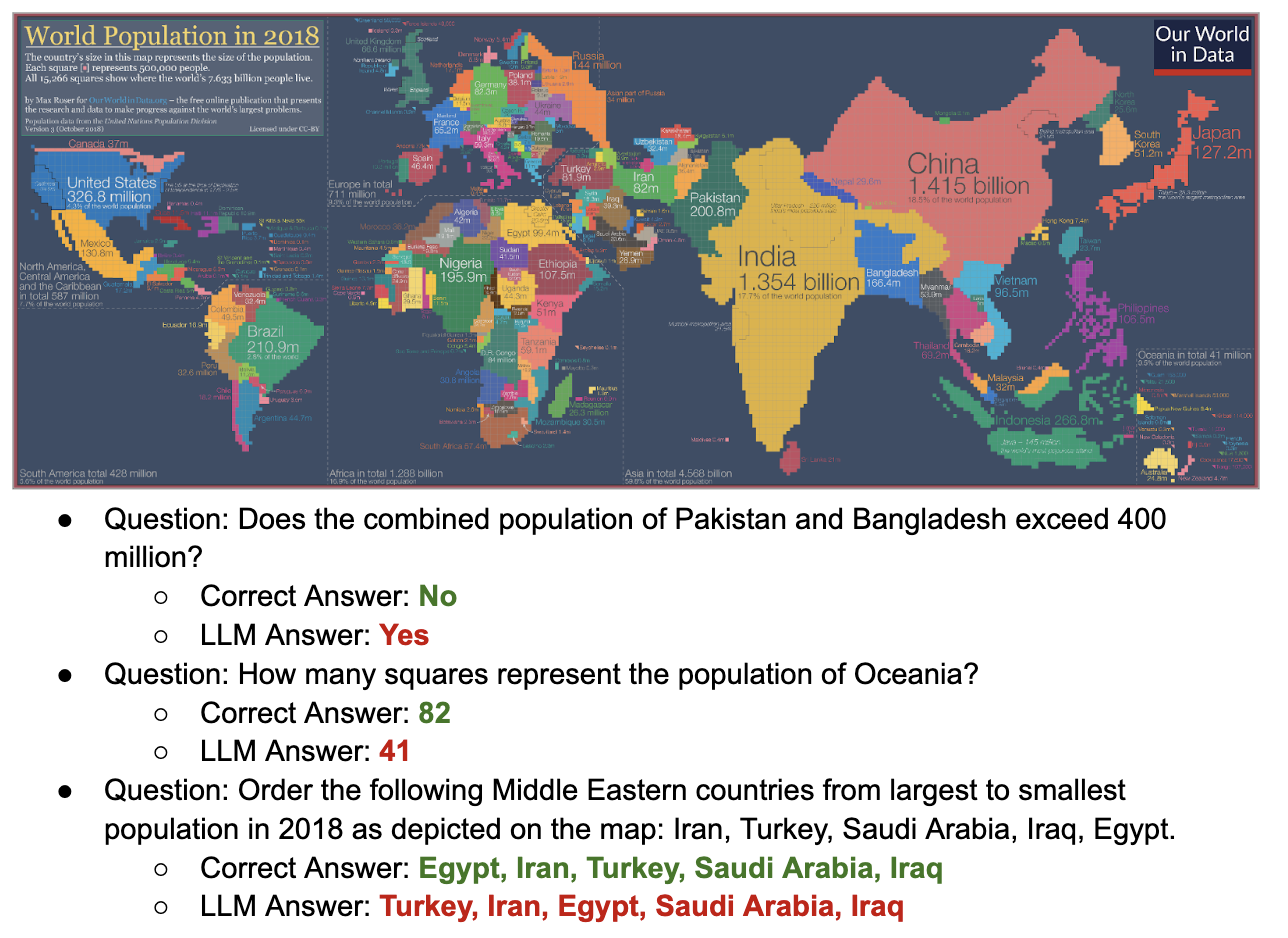}
    \caption{Sample QA for cartogram map type}
\end{figure}

\begin{figure}[!htb]
    \centering
    \includegraphics[width=1.0\linewidth]{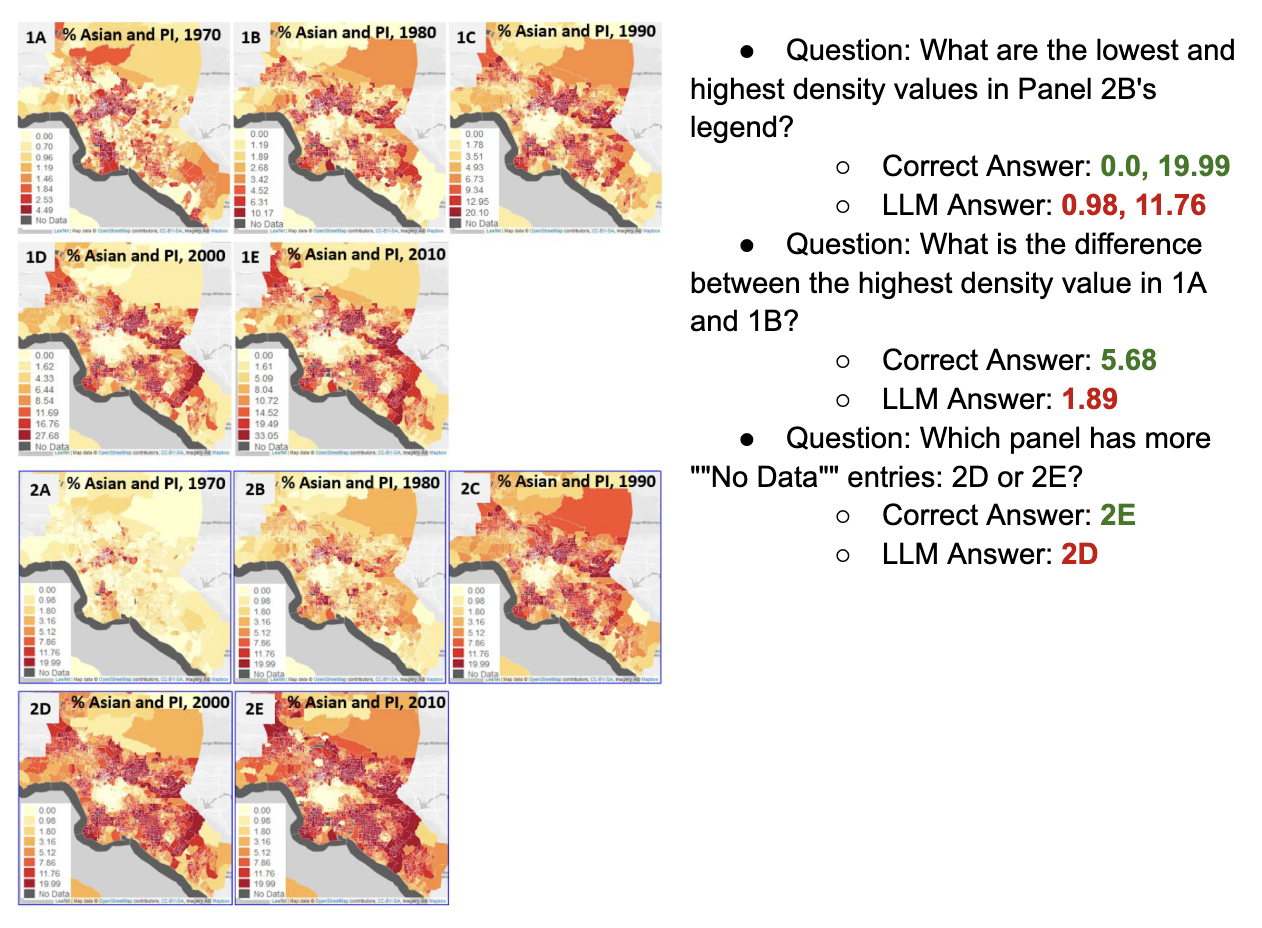}
    \caption{Sample QA for multiple map type}
\end{figure}

\onecolumn
\subsection{Custom Prompt for Question Answering}\label{ssec:prompt}

This subsection contains the full text of the custom prompt used to guide our AI agent's analysis of map data, as detailed in the main paper. The prompt provides a structured set of instructions and examples to ensure a consistent and logical approach to interpreting spatial information.

\begin{tcolorbox}[
    colback=white,
    colframe=black,
    title=\textbf{Custom Prompt},
    fonttitle=\bfseries\sffamily\large,
    halign title=center,
    boxsep=5pt,
    arc=4mm,
    boxrule=1pt,
    left=5pt,
    right=5pt
]
\raggedright 

You are an AI Agent with specialised knowledge in reading and understanding map data. Analyze the following map and using information from the steps and examples given below, answer the question.
\\
\textbf{Steps to follow:}
\begin{enumerate}
    \item Identify Map-Related Elements in the Question
    \item Locate the Identified Elements on the Map
    \item Apply Logical Reasoning
    \item Formulate a Concise Answer
\end{enumerate}

Based on your reasoning, arrive at a clear and accurate answer. Return only a word or phrase, as required—no explanation is needed.
If adequate data is not present, give answer as "no data".
If you have all the data and there is no answer to the question, give answer as "none". If it is a counting problem, give answer 0.
If you have all the data and it is not possible to answer the question, give answer "not possible".
\\
Assuming we are talking about a map with election results for USA. This map contains the voter breakdown across the United States, including the number of votes cast and the winning party in each state. Some examples of questions and their answers are as follows:
\\
\textbf{Question:} Count the number of states on the west coast where Democrats won.\\
\textbf{Answer:} 3

\textbf{Question:} Based on the information given in the map, who won the election, Democrats or Republicans?\\
\textbf{Answer:} Democrats

\textbf{Question:} Based on the information given in the map, if both Democrats and Republicans win 25 states each, do we have more blue states or red states?\\
\textbf{Answer:} neither

\textbf{Question:} List the top 4 states in terms of seats where the republicans won\\
\textbf{Answer:} Texas, Georgia, Missouri, Tennessee

\textbf{Question:} Rank these states in ascending order of seats - kansas, south carolina, nebraska, oklahoma, colorado, wisconsin\\
\textbf{Answer:} nebraska, kansas, oklahoma, south carolina, colorado, wisconsin

\textbf{Question:} Based on reasoning, Answer the following:
Montana : Wyoming :: North Dakota : ?\\
\textbf{Answer:} South Dakota

Now, Answer the Question below based on the information, instruction and examples above:
\end{tcolorbox}

\newpage 

\subsection{Annotator Instructions}\label{ssec:annotator_instructions}

The following instructions were provided to human annotators for creating the question-answer pairs in the \datasetName ~benchmark. The goal was to produce questions that require genuine geospatial reasoning and are difficult for large language models to answer correctly. 

\begin{tcolorbox}[
    colback=white,
    colframe=black,
    title=\textbf{Instructions},
    fonttitle=\bfseries\sffamily\large,
    halign title=center,
    boxsep=5pt,
    arc=4mm,
    boxrule=1pt,
    left=5pt,
    right=5pt
]
\label{sup:instructions} 
\textbf{Task Objective:} Your task is to create map-based questions along with answers based on the information provided in a map, ensuring that a human can easily answer them.

\begin{itemize}
    \item Provide objective questions grounded on the given map.
    \item Provide the correct answer to the question (single word or a few words).
    \item Include the answer provided by the LLM.
\end{itemize}

\textbf{Answers must be based on:}
\begin{itemize}
    \item The information presented in the map.
    \item General common understanding of how maps are read (spatial common sense).
\end{itemize}

\textbf{Instructions for Annotation:}
\begin{itemize}
    \item Use the "Add More Questions" button to get input spaces for question, correct answer, and answer by LLM.
    \item Once done creating all the questions, press the submit button. Your annotation would be saved by the portal.
\end{itemize}

\tcbline

\textbf{Preferred}
\begin{itemize}
    \item[\textbf{\ding{52}}] Analyze the map carefully to understand the information provided.
    \item[\textbf{\ding{52}}] Use only the information in the map to answer the question.
    \item[\textbf{\ding{52}}] Keep your questions simple and straightforward.
    \item[\textbf{\ding{52}}] Keep your answers concise (within a few words).
    \item[\textbf{\ding{52}}] Use common sense to provide additional context where necessary.
    \item[\textbf{\ding{52}}] Use proper grammar, spelling, and punctuation.
    \item[\textbf{\ding{52}}] Create at least 10 questions per map.
\end{itemize}

\tcbline

\textbf{Avoid}
\begin{itemize}
    \item[\textbf{\ding{56}}] Using outside knowledge not present in the map.
    \item[\textbf{\ding{56}}] Creating questions that are too long or difficult to understand.
    \item[\textbf{\ding{56}}] Creating ambiguous questions.
    \item[\textbf{\ding{56}}] Providing answers that are overly verbose or unclear.
    \item[\textbf{\ding{56}}] Providing subjective or ambiguous answers.
    \item[\textbf{\ding{56}}] Using abbreviations or acronyms not defined in the map.
    \item[\textbf{\ding{56}}] Creating more than three questions of the same type on a single map.
\end{itemize}

\tcbline

\textbf{Important Note}
Avoid using information that you may know if you believe it is not generally known.

The following examples clarify the instructions. Please review them carefully.

\tcbline

\href{https://docs.google.com/document/d/1nzCbCygIEdcZ62O8E3GZH5rB2ds5iFyM/edit}{[Link to Sample Annotation]}

\end{tcolorbox}

\end{document}